\documentclass[journal]{IEEEtran}

\usepackage{mathtools}
\usepackage[export]{adjustbox}
\usepackage{color}
\usepackage{graphics} 
\usepackage{epsfig} 
\usepackage{mathptmx} 
\usepackage{times} 
\usepackage{amsmath} 
\usepackage{amssymb}  
\usepackage{leftidx}
\newtheorem{theorem}{Theorem}
\newtheorem{remark}{Remark}
\usepackage[linesnumbered,ruled]{algorithm2e}
\usepackage{cite}[noadjust]
\usepackage[noend]{algpseudocode}

\newtheorem{assumption}{Assumption}
\title{Adaptive Sliding Mode Control for Vehicle Platoons with State-Dependent Friction Uncertainty}

\author{Rishabh Dev Yadav%
\thanks{Rishabh Dev Yadav is with the Robotics Research Center, International Institute of Information Technology Hyderabad, Hyderabad, India. Email: rishabh.yadav@research.iiit.ac.in.}%
\thanks{Thesis submitted in partial fulfillment of the requirements for the degree of Master of Science in Computer Science and Engineering by Research by RISHABH DEV YADAV (2021701030)}%
}

\begin{document}

\maketitle

\begin{abstract}
Multi-robot formation control has various applications in domains such as vehicle troops, platoons, payload transportation, and surveillance. Maintaining formation in a vehicle platoon requires designing a suitable control scheme that can tackle external disturbances and uncertain system parameters while maintaining a predefined safe distance between the robots. A crucial challenge in this context is dealing with the unknown/uncertain friction forces between wheels and the ground, which vary with changes in road surface, wear in tires, and speed of the vehicle. Although state-of-the-art adaptive controllers can handle a priori bounded uncertainties, they struggle with accurately modeling and identifying frictional forces, which are often state-dependent and cannot be a priori bounded.

This thesis proposes a new adaptive sliding mode controller for wheeled mobile robot-based vehicle platoons that can handle the unknown and complex behavior of frictional forces without prior knowledge of their parameters and structures. The controller uses the adaptive sliding mode control techniques to regulate the platoon's speed and maintain a predefined inter-robot distance, even in the presence of external disturbances and uncertain system parameters. This approach involves a two-stage process: first, the kinematic controller calculates the desired velocities based on the desired trajectory; and second, the dynamics model generates the commands to achieve the desired motion. By separating the kinematics and dynamics of the robot, this approach can simplify the control problem and allow for more efficient and robust control of the wheeled mobile robot.

The stability of the closed-loop system employing both the proposed controllers are studied analytically via Lyapunov theory. The effectiveness of the proposed controller is demonstrated through simulations using Gazebo, a popular robot simulation tool. The simulations show that the proposed controller outperforms existing state-of-the-art controllers in terms of stability, convergence, and robustness to changes in frictional forces. The simulations also demonstrate the controller's ability to maintain formation under various road conditions, including slopes, curves, and rough terrain.
\end{abstract}


\section{Introduction} \label{ch:intro}

The field of robotics has captured people's imaginations due to its potential for machines to replace humans in everyday tasks. One of the main benefits of robotics is the inefficiency of humans in performing repetitive tasks. Machines can perform these tasks without constraints such as boredom or distraction. Additionally, robots can transcend the mechanical limitations of humans, making them ideal for tasks that are physically demanding or hazardous. 

The development of robots has followed a similar path to the evolution of living beings. Initially, robots were designed for specific operations, and humanoid fantasies were seen as impractical with the technology available at the time. Robots came in various mechanical forms with task-specific features similar to living creatures. One significant step in the evolution of robotics was the development of autonomous systems that could perform tasks without human control. This development led to the need for a new science stream called control, which is essential for the functioning of autonomous systems.

One of the most significant applications of autonomous systems is in the field of autonomous vehicles. These vehicles have been brought to life through advances in control techniques and are used for research and commercial purposes. Autonomous vehicles come in various forms, including cars, ships, and quadrotors. While the use cases for aerial and marine robots are limited due to safety constraints, road vehicles equipped with autopilot features are already in use, and major automobile manufacturers are shifting their focus towards autonomous vehicles.

The control of autonomous vehicles encompasses a broad range of activities, from high-level control that selects an optimal trajectory to low-level control that determines appropriate actuator inputs. Robotics and control engineering are two closely related fields that have played an instrumental role in shaping our modern world. Robotics involves the design, construction, operation, and use of robots to perform a wide range of tasks, from manufacturing and assembly to exploration and space travel. Control engineering, on the other hand, focuses on designing and implementing control systems to manage and regulate complex processes, ensuring that they are efficient, safe, and reliable.

The integration of robotics and control engineering has led to many significant advancements in various industries. For example, in manufacturing, robots and automation have transformed production lines, allowing for faster and more efficient assembly of products. Robots can perform tasks that are repetitive, dangerous, or require high precision, which reduces the risk of injury and improves quality control. Control engineering is used to optimize and manage these complex systems, ensuring that they operate efficiently and safely.

Parts of this work were previously presented in \cite{yadav2021adaptive}; the present manuscript provides a consolidated and expanded exposition with complete derivations and simulations

\section{Autonomous Vehicle Platoon}

\begin{figure}[h]
\begin{center}
    \includegraphics[scale=0.45]{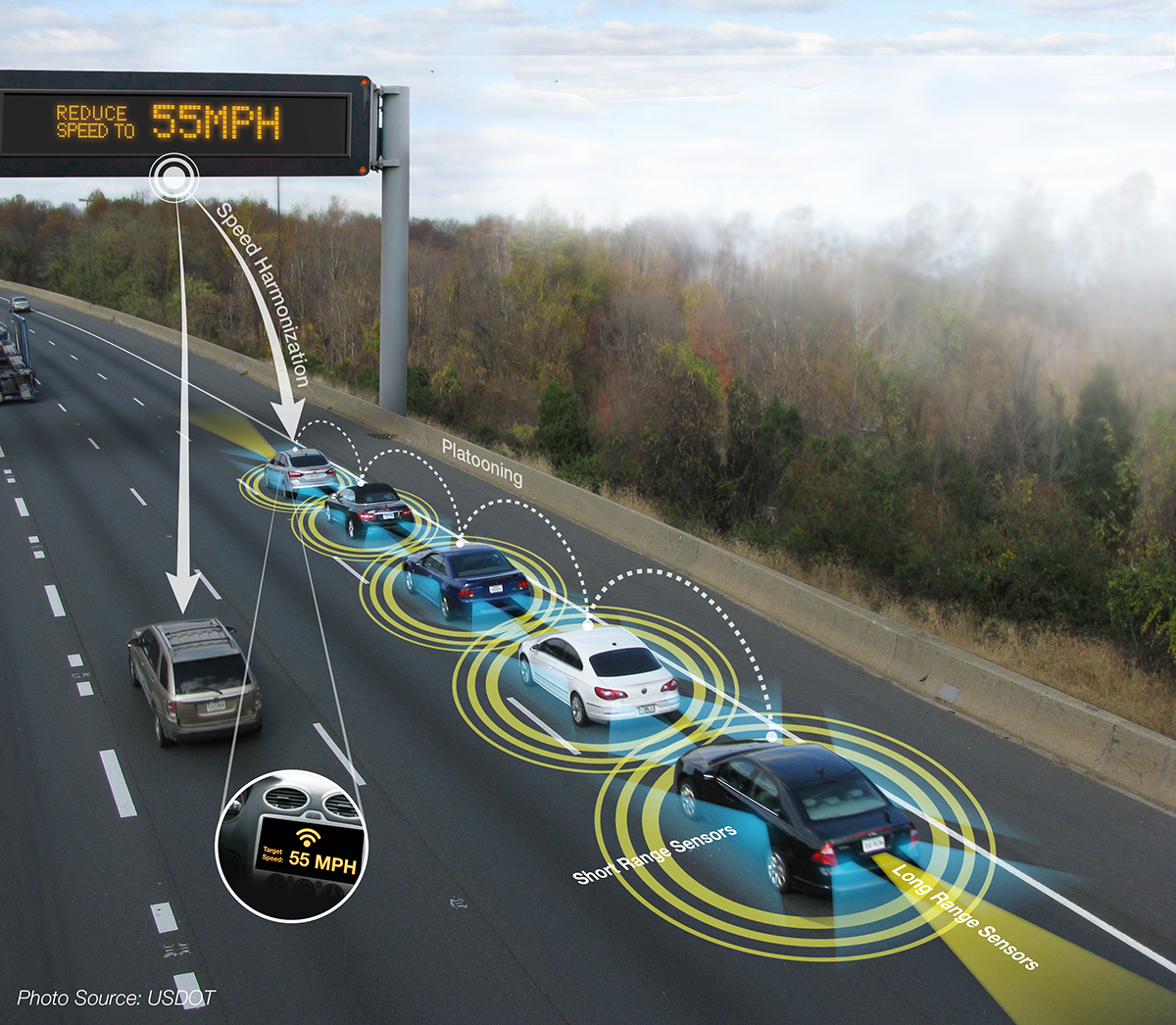}
    \caption{\small Platoonig. Source: Wikipedia}
    \label{robot_model}
\end{center}
\vspace{-4mm}
\end{figure}

Multi-wheeled mobile robot platoons are a type of robotic system that has the potential to revolutionize the way goods are transported in logistics and distribution networks. These platoons consist of a group of autonomous vehicles that can operate together in a coordinated manner, following a lead vehicle or a predetermined path. This technology is being developed by various research institutions and companies around the world, with the aim of reducing transportation costs and increasing efficiency.

The benefits of using multi-wheeled mobile robot platoons are numerous. First and foremost, they can increase transportation efficiency and reduce costs. By coordinating the movements of several vehicles, platoons can travel at a constant speed and maintain a safe following distance, reducing the need for drivers and optimizing fuel consumption. This can lead to significant savings for logistics and distribution companies.

Another benefit of multi-wheeled mobile robot platoons is their potential to reduce traffic congestion and improve safety on roads. By coordinating their movements, these platoons can reduce the number of vehicles on the road, decreasing the likelihood of accidents and reducing travel times for other drivers.

Additionally, these platoons can operate autonomously, eliminating the need for human drivers. This can lead to increased safety, as there is less risk of accidents caused by human error. It can also lead to cost savings for logistics and distribution companies, as they will not need to pay for drivers' salaries, benefits, and training.

\section{System Representation}
The main components involved in modeling a robotic system, namely, representing the model, defining reference frames with state variables, and developing kinematic and dynamic dynamic equations. Although there are various approaches to modeling dynamics in each of these sections, the focus will be on the method employed in this work.

\subsection{Reference Frames}
In order to describe the position of the robot Inertial Coordinate System (${X_I, Y_I}$) and Robot Coordinate System (${X_r, Y_r}$) frames are shown in Figure (\ref{robot_model}). 
The Inertial Coordinate System is a global frame which remains fixed in the environment or plane where the wheeled mobile robot (WMR) operates. It serves as the benchmark for reference. While, the Robot Coordinate System is local frame which is affixed to the mobile robot and moves along with it.

The position of any point on the
robot can be defined in the robot frame and the inertial frame as $X^r = [x^r, y^r, \theta^r]^T$ and $X^I = [x^I, y^I, \theta^I]^T$ respectively.
The important issue that needs to be explained at this stage is the
relation between these two frames. Then, the two coordinates are related by the following
transformation
\begin{align}
X^I = R(\theta) X^r
\end{align}
\begin{center}
    where $R(\theta) = \begin{bmatrix}
    cos(\theta) & -sin(\theta) & 0 \\
    sin(\theta) & cos(\theta) & 0 \\
    0 & 0 & 1
    \end{bmatrix}
            $
\end{center}

The robot position and orientation in the Inertial Frame can be defined as $q^I = [x_a, y_a, \theta]^T$

\subsection{Kinematic Model}
\begin{figure}[h]
\begin{center}
    \includegraphics[scale=0.45]{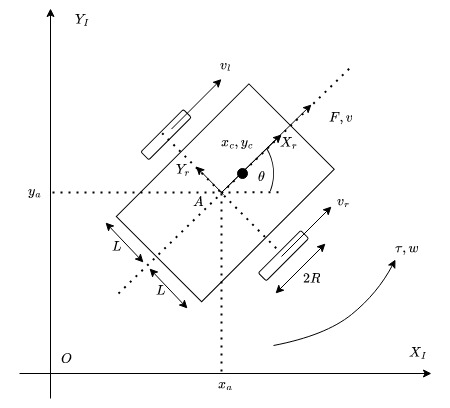}
    \caption{\small Schematic of a two-wheeled nonholonomic mobile robot}
    \label{robot_model}
\end{center}
\vspace{-4mm}
\end{figure}

Kinematic modeling involves analyzing the movement of mechanical systems while disregarding the forces that influence the movement. In the case of the mobile robot, the primary goal of kinematic modeling is to express the robot's velocities in relation to the velocities of its driving wheels and the robot's geometric characteristics.

Let $v_r$ and $v_l$ denote the right and left wheel linear speed respectively.
\begin{subequations}
\begin{gather}
v_r  = R\dot{\phi}_r \\
v_l  = R\dot{\phi}_l 
\end{gather}
\label{}
\end{subequations}
where $\phi_r$ and $\phi_l$ are angular speed of right and left wheel respectively; $R$ is wheel radius.
The linear velocity $v$ of the robot in the Robot Frame is the average of the linear velocities of the two wheels,
\begin{equation}
\label{v}
    v = \frac{v_r + v_l}{2} = R\frac{\dot{\phi}_r  + \dot{\phi}_l }{2}
\end{equation}
and the angular velocity of the robot is
\begin{equation}
\label{w}
    w = \frac{v_r - v_l}{2L} = R\frac{\dot{\phi}_r  - \dot{\phi}_l }{2}
\end{equation}
where $2L$ denote the width of the robot (cf. Fig. \ref{robot_model}).
The velocities in the robot frame can now be represented
in terms of the center-point A velocities in the robot frame as follows:
\begin{align}
\dot{x^r_a}  = v; ~~~~
\dot{y^r_a}  = 0; ~~~~
\dot{\theta}  = w
\end{align}

The robot velocities in the inertial frame can be written as 
\begin{equation}
\dot{q}^I =
    \begin{bmatrix}
    \dot{x}_a \\ \dot{y}_a \\ \dot{\theta} 
    \end{bmatrix} = 
    R(\theta)
    \begin{bmatrix}
    \frac{R}{2} & \frac{R}{2}\\
    0 & 0\\
     \frac{R}{2b} & -\frac{R}{2b}
    \end{bmatrix}
    \begin{bmatrix}
    \dot{\phi}_r \\ \dot{\phi}_l
    \end{bmatrix}
\end{equation}

For each robot in the platoon (cf. Fig. \ref{robot_model}), $ q = [x, y, \theta]^T$ represents the generalized state with $(x, y)$ is the robot position in the global inertial frame, and $\theta$ being its heading angle (yaw); $ u = [v, \omega]^T$ is the control input where $v$ is the linear velocity and $\omega$ is the angular velocity of the robot. The typical kinematic model of such system is given by 

\begin{equation}
    \dot{q} =  \begin{bmatrix} \cos(\theta) & 0\\
                            \sin(\theta) & 0\\
                            0 & 1 
                \end{bmatrix}u. \label{kin}
\end{equation}

\subsection{Dynamic model}
The field of dynamics involves examining how mechanical systems move while accounting for the various forces that impact their motion. This differs from kinematics, which does not take these forces into account. In order to analyze the motion of the mobile robot and develop motion control algorithms, it is crucial to have a dynamic model of the system.

Deriving system equations of motion via Lagrangian method rely on the following equation:
\begin{equation}
    \frac{d}{dt} (\frac{\partial L }{\partial \dot{q}_i}) -  \frac{\partial L}{\partial {q_i}} = F - \Gamma^T(q) \gamma
\label{ELeqtn}
\end{equation}
where $L = T-P$ is the Lagrangian function; $T$ and $P$ represents kinetic energy and potential energy of system respectively; $q_i$ are generalized coordinate; $F $ is the generalized force vector.However, since the robot is moving in the $(X_I, Y_I)$ plane, the potential energy of the robot is considered to be zero i.e. $P = 0$.

A non-holonomic differential drive robot with $n$ generalized coordinates ($q_1 ,q_2 ,...,q_n$) and subject to constraints can be described by the following equations of motion:
\begin{equation}
    M(q)\ddot{q} + V(q,\dot{q})\dot{q} + F(\dot{q}) + G(q) + \tau_d = B(q) {\tau_u} - \Gamma^T(q)\gamma
\label{generalDynamicsEq}
\end{equation}
where $ M(q)$ an $n \times n$ symmetric positive definite inertia matrix, $V(q,\dot{q})$ is
the centripetal and coriolis matrix, $F(\dot{q})$ is the surface friction matrix,
$G(q)$ is the gravitational vector, $\tau_d$ is the vector of bounded unknown
disturbances including unstructured unmodeled dynamics, $B(q)$ is the
input matrix, $\tau_u$ is the input vector, $\Gamma$ is the constraints
matrix, $\gamma$ is the vector of Lagrange multipliers associated with the
constraints.

Ignoring the distance between centre of mass of the robot and origin of the local coordinate frame attached to the mobile robot, the standard dynamic model of the system is given by
\begin{subequations}\label{dyn}
\begin{align}
m\dot{v} + f_v + d_v &= F,~~\text{with}~f_v =f_r + f_l, \label{dyn_a}\\
J\dot{\omega} + f_w + d_w &= \tau,~~\text{with}~f_w = f_r + f_l. \label{dyn_b}
\end{align}            
\end{subequations}
Here $m$ and $J$ denote the mass and the moment of inertia of the robot; $d_v$ and $d_w$ represent bounded external disturbances; $f_v$ and $f_w$ are the frictional force and frictional torque respectively generated at the right ($f_r$) and left ($f_l$) wheel; $(F, \tau )$ denote the control input.

Let $\tau_r$ and $\tau_l$ be the right and left wheel torque respectively and $R$ be the radius of the wheel. Then, the following holds 
\begin{equation}
    \begin{bmatrix} F \\
    \tau\\
                \end{bmatrix}
             =  \frac{1}{R}\begin{bmatrix} (\tau_r + \tau_l) \\
    (\tau_r - \tau_l)L
    \end{bmatrix}. 
\label{wheeltorque}
\end{equation}

\subsection{Friction Model}
Friction is the force that opposes motion between two surfaces in contact. Two common types of friction models are Coulomb friction and viscous friction. While both models describe the behavior of friction, they differ in their underlying assumptions and mathematical expressions.

Coulomb friction, also known as dry friction or static friction, is a simple model that assumes that the friction force between two surfaces in contact is proportional to the normal force pressing the surfaces together. The model further assumes that the friction force is independent of the sliding speed and that it only acts when the surfaces are in contact and stationary relative to each other. Once the surfaces start moving relative to each other, the friction force decreases and becomes proportional to the kinetic friction coefficient, which is typically lower than the static friction coefficient.

Viscous friction, also known as dynamic or fluid friction, is a more complex model that assumes that the friction force is proportional to the velocity of the surfaces relative to each other. This model is often used to describe the behavior of fluids, such as air or liquids, but can also be applied to solid materials. The viscous friction coefficient depends on the viscosity of the fluid or the deformation properties of the solid material and can vary with temperature and other factors.

\begin{figure}[h]
\begin{center}
    \includegraphics[scale=0.45]{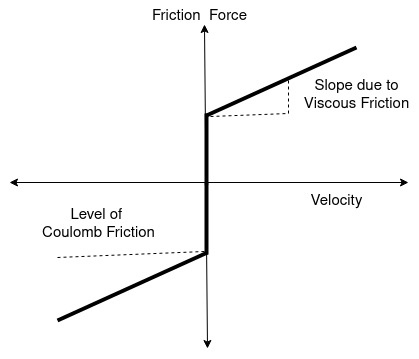}
    \caption{\small Coulomb + Viscous friction model}
    \label{friction_model_pic}
\end{center}
\vspace{-4mm}
\end{figure}

The mathematical expressions for Coulomb and viscous friction are also different. Coulomb friction is typically expressed as a constant static friction coefficient and a lower kinetic friction coefficient, while viscous friction is expressed as a linear function of the velocity between the surfaces.

While both models have their limitations, they are useful for understanding and predicting the behavior of friction in different scenarios. The general friction model for a moving body considering the static Coulomb friction ($f_k$) and the viscous friction ($f_c$) can be represented as
$f_{r} = f_{k} + f_c $ (cf. \ref{friction_model_pic}). The friction forces $f_v$ and $f_w$ in (\ref{dyn_a}) and (\ref{dyn_b}) can be rewritten as
\begin{subequations}
\begin{align}
    f_v &= (f_{kr} + f_{cr} v_r) + (f_{kl} + f_{cl} v_l), \label{friction_eq_a}\\
    f_w &= ((f_{kr} + f_{cr} v_r) - (f_{kl} + f_{cl} v_l))L \label{friction_eq_b}
\end{align}
\label{friction_eq}
\end{subequations}
where $f_{ki}$ and $f_{ci}$ are coulomb and viscous friction respectively for wheel having speed $v_i$ with $i= r,l$ as in (\ref{v}) and (\ref{w}).

\section{Control of multiple mobile robots}

\subsection{Motivation}
Cooperative mobile multi-robot systems have gained significant attention due to their advantages over single-robot systems, such as better efficiency, high tolerance, redundancy, and manoeuvrability \cite{darmanin2017review,arai2002advances}. These multi-robot systems have a broad range of applications, including search and rescue, exploration, navigation, security and surveillance, precision agriculture, and payload transportation \cite{ismail2018survey, cai2012survey, cortes2017coordinated,soni2018formation}. One specific application is the platoon of autonomous vehicles, where multiple agents/robots/vehicles follow a common path in a shared environment. The control objective of such formation control is often decentralized \cite{antonelli2006kinematic} , where the aim is to maintain a desired line-of-sight range between each vehicle and its predecessor while proceeding along a given trajectory \cite{dai2017platoon}. However, controlling these systems under parametric uncertainties and unmodelled dynamics is still a significant challenge and an open problem.

\subsection{Related Works}
When it comes to autonomous platooning, the accuracy of formation control is heavily influenced by vehicle dynamics and external disturbances, particularly at higher speeds. The use of a simple Coulomb friction model is inadequate to capture the complex relationship between tyre wear and friction forces, as evidenced by previous studies \cite{berntorp2019steering,iurian2005identification, vcerkala2014mobile}. However, accurately parameterizing this phenomenon in real-world situations is challenging, if not impossible, due to the variability of friction forces caused by factors such as changes in payload quantity, road conditions, and tyre distortion.

Modern robotics research increasingly emphasizes reliable operation under uncertainties, constraints, and complex system dynamics. To address these challenges, researchers have explored a range of methods including adaptive control, robust system design, and advanced planning strategies. Such approaches play a crucial role in improving the performance and dependability of robotic systems in practical applications. Representative studies in this area are presented in \cite{ganguly2021efficient, sankaranarayanan2022robustifying, ganguly2022robust, harithas2022cco, suraj2022introducing, dantu2022adaptive, dantu2023adaptive, gupta2024adaptive, yadav2024modular, dantu2024adaptive, yadav2025integrated, sharma2025impedance}.

Advanced control strategies such as interpolating control \cite{tuchner2017vehicle}, distributed formation control \cite{xu2019distributed}, intervehicle distance control \cite{zhang2020analysis}, and robust control \cite{zhao2018robust} have been developed to handle the dynamics of vehicles for both longitudinal and lateral control in automatic platoon formation. However, these strategies require prior knowledge of system parameters. To address this limitation, two-stage tracking controllers have been used for wheeled mobile robots \cite{chen2009design, asif2014adaptive, koubaa2018adaptive}, combining a kinematic controller with an adaptive sliding mode controller (ASMC). However, these controllers can only handle bounded uncertainties, and frictional forces, which are state-dependent, do not satisfy this requirement \cite{roy2019overcoming, utkin2013adaptive}.

Therefore, there is a need for an adaptive control solution that can handle state-dependent unknown dynamics without prior knowledge for a multi-robot platoon system. Current research in this area has been focused on using a combination of a kinematic controller and ASMC for tracking control. However, these methods have been limited in their ability to handle unbounded uncertainties, which can arise due to the state-dependent nature of frictional forces.
To overcome this limitation, new research is needed to develop adaptive control methods that can handle state-dependent uncertainties in multi-robot platoon systems. This research should focus on developing controllers that can adapt to changing conditions in real-time, without requiring prior knowledge of system parameters. In addition, these controllers should be designed to be robust to unbounded uncertainties, such as those arising from frictional forces.

\subsection{Contribution}
Based on the limitations of current control methods for multi-robot platoon systems, a new framework ASMC is proposed. This framework does not require prior knowledge of system dynamics, including inertia parameters, frictional forces, and external disturbances. The stability of the closed-loop system is analyzed using the Lyapunov method, and simulation results show that the proposed method outperforms current state-of-the-art control methods for multi-robot platoon formation control.

\section{Organization of the Thesis}
The thesis is organized into four chapters. A brief summary of each chapter is mentioned below.

\begin{itemize}
    \item \textbf{Chapter 1:} This introductory chapter gives an overview of wheeled robotics, kinematic and dynamic modelling and state-of-the-art control strategies. It briefly describes the motivation for this research, the problem orientation, the pertaining gaps in the literature, the main contributions and an outline of the thesis.
    
    \item \textbf{Chapter 2:} The chapter expalins a new adaptive sliding mode controller for WMR-based vehicle platoons that can handle unknown and complex frictional forces. The controller maintains a predefined inter-robot distance and regulates the platoon's speed despite external disturbances and uncertain system parameters. The approach involves a two-stage process of kinematic and dynamic controllers to achieve the desired motion. This allows for more efficient and robust control of the mobile robot. The stability of the closed-loop system using the proposed controllers is studied using Lyapunov theory.
    
    \item \textbf{Chapter 3:} In this chapter the proposed controller's effectiveness was demonstrated using Gazebo simulations results, showing it outperformed state-of-the-art controllers in terms of stability, convergence, and robustness to frictional force changes. The performance is compare via error plots and root mean-squared (RMS) error.
    
    \item \textbf{Chapter 4:} This chapter concludes the thesis by summarizing the various contributions brought out by this thesis.
    
\end{itemize}


In cases where the system parameters are unknown, state-of-art control laws have been employed to handle a priori bounded uncertainties. Unfortunately, frictional forces generally do not adhere to such uncertainty settings due to their state-dependent nature. Prior studies have been unable to handle state-dependent unknown dynamics without prior knowledge, leaving a gap in the field. Therefore, a solution for multi-robot platoon systems that can handle state-dependent unknown dynamics without prior knowledge is still lacking.

Toward this direction, the proposed adaptive control solution has the following major contributions:
\begin{itemize}
    \item The study introduces an ASMC framework to address state-dependent dynamic factors such as frictional and inertial forces, as well as external disturbances, in each vehicle of a platoon.
    
    \item The closed-loop stability of the system is analysed via Lyapunov-based method and comparative simulation results suggest significant improvement in tracking accuracy of the proposed scheme compared to the state of the art.
\end{itemize}

\section{CONTROL
FORMULATION}\label{sec:controller_design}
The platoon control problem involves multiple vehicles, each with their own trajectory to follow. Except for the leading vehicle, each follower calculates its next waypoint based on the position of the vehicle in front of it. The focus of this work is on controlling each vehicle, which is considered a nonholonomic WMR, to follow the desired path rather than planning the path for each robot.

Controlling a nonholonomic WMR effectively requires considering both kinematic and dynamic model-based controllers \cite{chen2009design, asif2014adaptive, koubaa2018adaptive}. In this approach, the linear and angular velocity derived from the kinematic controller are used as the desired trajectory in the dynamic model. However, since parametric uncertainty can only be captured in the dynamic model, the novelty of this work lies in designing an ASMC for the dynamic model while using the standard kinematic controller as in previous works \cite{fierro1997control,garcia2017tracking,liu2020adaptive}.

\subsection{Kinematic Control Design}

The kinematic control design objective is to follow a time-varying reference trajectory 
    $q_r(t) = [x_r(t),   y_r(t),  \theta_r(t)]^T$.

The following standard assumption is made:
\begin{assumption}
\label{assum1}
The desired trajectories $x_r(t)$ and $y_r(t)$ are designed to
be sufficiently smooth and bounded.
\end{assumption}

The posture tracking error of the mobile robot $q_e(t)$ is defined as
\begin{equation}
    q_e(t) = [e_1(t),   e_2(t),  e_3(t)]^T
\end{equation}
where 
\begin{equation}
  \begin{bmatrix} {e_1(t)} \\
    {e_2(t)}\\
    {e_3(t)}\\
                \end{bmatrix}  =
      \begin{bmatrix} \cos\theta(t) &  \sin\theta(t) & 0\\
    -\sin\theta(t) & \cos\theta(t) & 0\\
                            0 & 0 & 1
            \end{bmatrix} 
            \begin{bmatrix} x_r(t) - x(t)\\
    y_r(t) - y(t)\\
    \theta_r(t) - \theta(t)\\
                \end{bmatrix}. \label{e}
   \end{equation}       
In order to be more concise, we will eliminate the time-related aspects of the functions whenever feasible. Using \eqref{kin}, the time derivative of 
(\ref{e}) leads to
 \begin{equation}
    \begin{bmatrix} \dot{e}_1 \\
    \dot{e}_2\\
    \dot{e}_3\\
                \end{bmatrix}
             = v \begin{bmatrix} -1 \\
    0\\
    0\\
                \end{bmatrix}  + 
            \omega \begin{bmatrix} {e_2} \\
    -{e_1}\\
    -1\\
                \end{bmatrix} +
        \begin{bmatrix}  v_d \cos(e_3)\\
    v_d \sin(e_3)\\
    \omega_d\\
                \end{bmatrix},
    \end{equation}
where $ u_d = [v_d, \omega_d]^T$ denotes the reference (desired) time-varying linear and angular velocity.

Following \cite{fierro1997control,garcia2017tracking,liu2020adaptive}, we used the following backstepping method based kinematic tracking control law
\begin{equation}
    \begin{bmatrix} v_c \\
    \omega_c\\
                \end{bmatrix}
             =  \begin{bmatrix} v_d \cos(e_3) + k_1 e_1\\
    \omega_d + k_2 v_d e_2 + k_3 v_d \sin(e_3)\\
    \end{bmatrix} 
\end{equation}
where $k_1$, $k_2$ and $k_3$ are positive design constants. 

As mentioned earlier, the main contribution of the work lies in designing the dynamic controller and the corresponding control problem is discussed subsequently.

\subsection{Proposed Dynamic Controller Design}
\begin{remark}[State-dependent forces]
It is crucial to note that viscous friction being proportional to the velocity $v$ of the vehicle, the friction forces are state-dependent and thus, cannot be bounded a priori \cite{roy2020towards, roy2020adaptive,roy2021adaptive, shukla2021robust}. Such consideration segregates this work from the state-of-the-art adaptive solutions \cite{chen2009design, asif2014adaptive, koubaa2018adaptive} relying on a priori bounded dynamical forces.  
\end{remark}

With this observation, we present the following assumption on system dynamics uncertainty, which acts as a control design challenge.

\begin{assumption}[Uncertainty setting]
\label{assum2}
The system dynamics terms $m, J, d_v, d_w$ and their bounds are unknown for control design.
\end{assumption}

Following the system dynamics structure \eqref{dyn}, the proposed control framework is divided into two parts, namely, force control and torque control as detailed in the following two subsections. It is noteworthy that the co-design approach of force and torque control are not independent; rather they are interconnected via the uncertainty structures (cf. (\ref{varphiv}) and (\ref{varphiw})) and thereby to be designed simultaneously.

Let us define the linear velocity tracking error $e_v$ and the angular velocity tracking error $e_\omega$ as
\begin{equation} \label{err}
   e_v \triangleq v - v_c, ~~ e_\omega \triangleq \omega - \omega_c.
\end{equation}

\subsection{Force Control}
Let the sliding variable be designed as
\begin{equation}
    s_v(t) = e_v(t) +  \phi_v \int_{0}^{t} e_v(\psi) \,d\psi ,
\label{sliding_funv}
\end{equation}
where $\phi_v$ is a positive design scalar. Multiplying the
time derivative of (\ref{sliding_funv}) by $m$ and using (\ref{dyn}) gives
\begin{equation}
    m\dot{s}_v = m(\dot{v} - \dot{v}_c + \phi_v e_v) = F + \epsilon_v ,
\label{msvdot}
\end{equation}
where $\epsilon_v \triangleq  -( d_v + m\dot{v}_c + f_v - m \phi_v e_v )$ represents the overall uncertainty in force dynamics with its upper bound structure computed as
\begin{equation}
    |\epsilon_v|  \leq \overline{d_v} + m (|\dot{v}_c| + |\phi_v| |e_v|) + |f_v| \label{up_f},
\end{equation}
where $|{d_v}| \leq \overline {d_v}$. Substituting the relations $v_r = \frac{2v + \omega L}{2}$ and $v_l = \frac{2v - \omega L}{2}$ from (\ref{v}) and (\ref{w}) into (\ref{friction_eq}), the upper bound structure (\ref{up_f}) can be simplified to
\begin{align}
    |\epsilon_v| \leq & \overline{d_v} + m (|\dot{v}_c| + |\phi_v| |e_v|) + |f_{kr}| + |f_{kl}| + \nonumber \\
    & |v| (|f_{cr}| + |f_{cl}|) + 
    (L/2)  |\omega|(|f_{cr}| + |f_{cl}|).
    \label{absphiv2}
\end{align}
Further, let us define $\xi_v \triangleq [e_v, \int_{0}^{t} e_v(\psi) \,d\psi]^T$ and $\xi_\omega \triangleq [e_\omega, \int_{0}^{t} e_\omega(\psi) \,d\psi]^T$. Then, using the inequalities $|\xi_v| \geq |e_v|$ and $|\xi_\omega| \geq |e_\omega|$ and substituting $v = e_v + v_c$ and $\omega = e_\omega + \omega_c$ from (\ref{err}) into (\ref{absphiv2}) one obtains 
\begin{align}
 |\epsilon_v| \leq & {K}^*_{v0} + {K}^*_{v1} |\xi_v| + {K}^*_{\omega 2} |\xi_\omega| 
\label{varphiv} \\
 \text{where}~   {K}^*_{v0} \triangleq & \overline{d_v} + m |\dot{v}_c| + |f_{kr}| + |f_{kl}| + |v_c|(|f_{cr}| + |f_{cl}|) \nonumber \\
    & + (L/2) |\omega_c| (|f_{cr}| + |f_{cl}|), \nonumber \\
    {K}^*_{v1} \triangleq& m |\phi_v| + (|f_{cr}| + |f_{cl}|), ~ {K}^*_{\omega 2} \triangleq (L/2) (|f_{cr}| + |f_{cl}|) \nonumber
\end{align}
are unknown scalars.

The force control law is designed as
\begin{subequations}
\begin{gather}
   F(t) = - \Lambda_v s_v(t) - \rho_v(t) sgn(s_v(t)), \\
   \rho_v(t) = \hat{K}_{v0}(t) + \hat{K}_{v1} |\xi_v| +  \hat{K}_{\omega 2} |\xi_\omega|,
\end{gather}
\label{force_law}
\end{subequations}
where $\Lambda_v$ is a positive scalar gain and $(\hat{K}_{v0}(t), \hat{K}_{v1}(t) \hat{K}_{\omega 2}(t))$ are estimates of $({K}^*_{v0}, {K}^*_{v1}, {K}^*_{\omega 2}  )$ obtained via the following adaptive laws
\begin{subequations}
\begin{align}
     \dot{\hat{K}}_{v0}(t) = |s_v(t)|  - \alpha_{v0} \hat{K}_{v0}(t), \hat{K}_{v0}(0) > 0, \\
    \dot{\hat{K}}_{v1}(t) = |s_v(t)| |\xi_v(t)| - \alpha_{v1} \hat{K}_{v1}(t), \hat{K}_{v1}(0) > 0, \\
    \dot{\hat{K}}_{\omega 2}(t) = |s_\omega(t)| |\xi_\omega(t)| - \alpha_{\omega 2} \hat{K}_{\omega 2}(t), \hat{K}_{\omega 2}(0) > 0,
\end{align}
\label{adaptivelaw_force}
\end{subequations}
where $\alpha_{v0}$, $\alpha_{v1}$, $\alpha_{\omega 2}$ are user-defined positive scalars.

\subsection{Torque Control}
For torque control, the sliding variable is designed as
\begin{equation}
    s_\omega(t) = e_\omega(t) +  \phi_\omega \int_{0}^{t} e_\omega(\psi) \,d\psi ,
\label{sliding_funw}
\end{equation}
where $\phi_\omega$ is a positive user-defined gain. Multiplying the
time derivative of (\ref{sliding_funw}) by $J$ and using (\ref{dyn}) yields
\begin{equation}
    J\dot{s_\omega} = J(\dot{\omega} - \dot{\omega_c} + \phi \omega_v) = \tau + \epsilon_\omega ,
\label{mswdot}
\end{equation}
where $\epsilon_\omega \triangleq  -( d_\omega + J\dot{\omega_c} + f_\omega - J \phi_\omega e_\omega)$ is the overall uncertainty for the torque dynamics and it satisfies the following upper bound structure 
\begin{equation}
    |\epsilon_\omega|  \leq \overline{d_\omega} + J (|\dot{\omega_c}| + |\phi_\omega| |e_\omega|) + |f_\omega| \label{un_w}
\end{equation}
where $|d_\omega| \leq \overline{d_\omega}$. Substituting $v_r$ and $v_l$ from (\ref{v}) and (\ref{w})  into (\ref{friction_eq}), (\ref{un_w}) is simplified to
\begin{align}
    |\epsilon_\omega| \leq & \overline{d_\omega} + J (|\dot{\omega_c}| + |\phi_\omega| |e_\omega|) + L(|f_{kr}| + |f_{kl}|) +  \nonumber \\
    & |v| L ( |f_{cr}| + |f_{cl}|) +  |\omega| (L^2/2)  (|f_{cr}| + |f_{cl}|)
    \label{absphiw2}.
\end{align}
Substituting $v = e_v + v_c$ and $\omega = e_\omega + \omega_c$ in (\ref{absphiw2}) and using the
inequalities $|\xi_v| \geq |e_v|$ and $|\xi_\omega| \geq |e_\omega|$ the following is obtained from (\ref{absphiw2})
\begin{align}
&|\epsilon_\omega| \leq  {K}^*_{\omega 0} + {K}^*_{ \omega1} |\xi_\omega| + {K}^*_{v2} |\xi_v|
\label{varphiw} \\
\text{where}~ {K}^*_{\omega 0} \triangleq & \overline{d_\omega} + J |\dot{\omega_c}| + L(|f_{kr}| + |f_{kl}|) + |v_c| L(|f_{cr}| + |f_{cl}|) \nonumber \\
 &+  |\omega_c| (L^2/2)  (|f_{cr}| + |f_{cl}|) , \nonumber \\
 {K}^*_{\omega 1} \triangleq & J |\phi_\omega| + (L^2/2)  (|f_{cr}| + |f_{cl}|),~  {K}^*_{v2} \triangleq  L(|f_{cr}| + |f_{cl}|) \nonumber
\end{align}
are unknown scalars.

The torque control law is designed as
\begin{subequations}
\begin{gather}
   \tau(t) = - \Lambda_\omega s_\omega(t) - \rho_\omega(t) sgn(s_\omega(t)), \\
   \rho_\omega(t) = \hat{K}_{\omega 0}(t) + \hat{K}_{\omega 1} |\xi_\omega| +  \hat{K}_{v 2} |\xi_v|,
\end{gather}
\label{torque_law}
\end{subequations}
where $\Lambda_\omega$ is a user-defined positive scalar gain. The adaptive gains $\hat{K}_{\omega 0}(t)$, $\hat{K}_{\omega 1}(t)$, $\hat{K}_{v 2}(t)$ are the estimates of ${K}^*_{\omega 0}, {K}^*_{\omega 1}, {K}^*_{v2}$ updated as
\begin{subequations}
\begin{align}
     \dot{\hat{K}}_{\omega0}(t) = |s_\omega(t)|  - \alpha_{\omega 0} \hat{K}_{\omega 0}(t), \hat{K}_{\omega 0}(0) > 0, \\
    \dot{\hat{K}}_{\omega 1}(t) = |s_\omega(t)| |\xi_\omega(t)| - \alpha_{\omega 1} \hat{K}_{\omega 1}(t), \hat{K}_{\omega 1}(0) > 0 ,\\
    \dot{\hat{K}}_{v 2}(t) = |s_v(t)| |\xi_v(t)| - \alpha_{v2} \hat{K}_{v2}(t), \hat{K}_{v2}(0) > 0 
\end{align}
\label{adaptivelaw_torque}
\end{subequations}
where $\alpha_{\omega 0}$, $\alpha_{\omega 1}$, $\alpha_{v2}$ are user-defined positive scalars.
\begin{remark}
The upper bound structures of $\epsilon_v$ and $\epsilon_w$ in (\ref{varphiv}) and (\ref{varphiw}), respectively, reveal that state-dependencies occur
inherently in the system uncertainties
via $\xi_v$ and $\xi_w$. Therefore, the gains $\rho_v$ and $\rho_w$ in (\ref{force_law}) and (\ref{torque_law}), respectively, are designed according to these state-dependent structures.
\end{remark}

\section{Stability Analysis of the Proposed Controller}
\begin{theorem}
Under the Assumptions 1 and 2, the closed-loop trajectories of (\ref{msvdot}) and (\ref{mswdot}) with control laws (\ref{force_law}) and (\ref{torque_law}), along with the adaptive laws (\ref{adaptivelaw_force}) and (\ref{adaptivelaw_torque}) are Uniformly Ultimately Bounded (UUB). 
\end{theorem}

The closed-loop stability analysis is carried out using the following Lyapunov function
\begin{align}
 V = & V_v + V_\omega,\label{lyp}\\
\text{where}~    V_v = & (1/2) \big [ms_v^2 +  (\hat{K}_{v0} - {K}^*_{v0})^2 + (\hat{K}_{v1} - {K}^*_{v1})^2    \nonumber \\
     &+ (\hat{K}_{\omega 2} - {K}^*_{\omega 2})^2 \big ]\nonumber,
\\
    V_w = & (1/2) \big [ Js_w^2 +  (\hat{K}_{w0} - {K}^*_{w0})^2 + (\hat{K}_{w1} - {K}^*_{w1})^2 \nonumber \\ 
    &+ (\hat{K}_{v2} - {K}^*_{v2})^2 \big ]\nonumber.
\end{align}

Using (\ref{msvdot}), the time derivative of $V_v$ yields
\begin{align}
    \dot{V}_v = & m s_v \dot{s_v} +(\hat{K}_{v0} - {K}^*_{v0})\dot{\hat{K}}_{v0} +  (\hat{K}_{v1} - {K}^*_{v1})\dot{\hat{K}}_{v1} \nonumber \\
    & +(\hat{K}_{w2} - {K}^*_{w2})\dot{\hat{K}}_{w2} \nonumber\\
    = & s_v(F + \epsilon_v) + (\hat{K}_{v0} - {K}^*_{v0})\dot{\hat{K}}_{v0} +  (\hat{K}_{v1} - {K}^*_{v1})\dot{\hat{K}}_{v1} \nonumber \\
     &+ (\hat{K}_{w2} - {K}^*_{w2})\dot{\hat{K}}_{w2}.
\label{Vvdot}
\end{align}
   
Using the control law (\ref{force_law}) and the upper bound from (\ref{varphiv})  we have
\begin{align} 
    \dot{V}_v =& s_v(- \Lambda_v s_v - \rho_v sgn(s_v)+\epsilon_v) + (\hat{K}_{v0} - {K}^*_{v0})\dot{\hat{K}}_{v0} \nonumber \\
    & + (\hat{K}_{v1} - {K}^*_{v1})\dot{\hat{K}}_{v1} +  (\hat{K}_{w2} - {K}^*_{w2})\dot{\hat{K}}_{w2} \nonumber\\
     = & - \Lambda_v s_v^2 - (\hat{K}_{v0} - {K}^*_{v0})(|s_v| - \dot{\hat{K}}_{v0}) \nonumber \\
    & - (\hat{K}_{v1} - {K}^*_{v1})(|s_v| |\xi_v| - \dot{\hat{K}}_{v1}) \nonumber \\
    &-     (\hat{K}_{w2} - {K}^*_{w2})(|s_w| |\xi_w| - \dot{\hat{K}}_{w2}).
\label{replacekv}
\end{align}
The adaptive laws in (\ref{adaptivelaw_force}) yield
\begin{align}
(\hat{K}_{v0} - {K}^*_{v0}) \dot{\hat{K}}_{v0} = & |s_v|(\hat{K}_{v0} - {K}^*_{v0}) + \alpha_{v0} \hat{K}_{v0} {K}^*_{v0} \nonumber \\ 
& -\alpha_{v0} \hat{K}_{v0}^2
\label{kv0} \\
(\hat{K}_{v1} - {K}^*_{v1}) \dot{\hat{K}}_{v1} = & |s_v|(\hat{K}_{v1} - {K}^*_{v1}) |\xi_v| + \alpha_{v1} \hat{K}_{v1} {K}^*_{v1} \nonumber \\
& -\alpha_{v1} \hat{K}_{v1}^2
\label{kv1} \\
(\hat{K}_{w2} - {K}^*_{w2}) \dot{\hat{K}}_{w2} = & |s_w|(\hat{K}_{w2} - {K}^*_{w2}) |\xi_w| + \alpha_{w2} \hat{K}_{w2} {K}^*_{w2} \nonumber \\ 
& -\alpha_{w2} \hat{K}_{w2}^2.
\label{kw2}
\end{align}
Substituting (\ref{kv0})-(\ref{kw2}) into (\ref{replacekv}) yields
\begin{align}
\dot{V}_v \leq & -\Lambda_v |s_v|^2 + (\alpha_{v0} \hat{K}_{v0} {K}^*_{v0} - \alpha_{v0} \hat{K}_{v0}^2) \nonumber \\ 
& + (\alpha_{v1} \hat{K}_{v1} {K}^*_{v1} - \alpha_{v1} \hat{K}_{v1}^2) 
 + (\alpha_{w2} \hat{K}_{w2} {K}^*_{w2} -  \alpha_{w2} \hat{K}_{w2}^2) \nonumber \\
 \leq & -\Lambda_v |s_v|^2 - (1/2) \alpha_{v0} ((\hat{K}_{v0} - {K}^*_{v0})^2 - {{K}^*_{v0}}^2) \nonumber \\ 
 & - (1/2) \alpha_{v1} ((\hat{K}_{v1} - {K}^*_{v1})^2 - {{K}^*_{v1}}^2) \nonumber \\
 & -
 (1/2) \alpha_{w2} ((\hat{K}_{w2} - {K}^*_{w2})^2 - {{K}^*_{w2}}^2).
 \label{Vvdotineq2}
\end{align}
Following the similar lines to obtain (\ref{Vvdotineq2}), one can also obtain $\dot{V}_w$ as the following  
\begin{align}
 \dot{V}_w \leq & -\Lambda_w |s_w|^2 - (1/2) \alpha_{w0} ((\hat{K}_{w0} - {K}^*_{w0})^2 - {{K}^*_{w0}}^2) \nonumber\\
 &- (1/2) \alpha_{w1} ((\hat{K}_{w1} - {K}^*_{w1})^2 - {{K}^*_{w1}}^2) \nonumber\\
 &- (1/2) \alpha_{v2} ((\hat{K}_{v2} - {K}^*_{v2})^2 - {{K}^*_{v2}}^2). \label{new1}
\end{align}
 Further, using the definitions of Lyapunov function as in (\ref{lyp}), (\ref{Vvdotineq2}) and (\ref{new1}) can be further simplified to
 \begin{align}
   &  \dot{V}_v  \leq  -\varrho_v V_v + \frac{1}{2}(\alpha_{v0} {{K}^*_{v0}}^2  + \alpha_{v1} {{K}^*_{v1}}^2 + \alpha_{w2} {{K}^*_{w2}}^2) ,
\label{Vvdotineq3}\\
 &    \dot{V}_w  \leq   -\varrho_w V_w + \frac{1}{2}(\alpha_{w0} {{K}^*_{w0}}^2  + \alpha_{w1} {{K}^*_{w1}}^2 + \alpha_{v2} {{K}^*_{v2}}^2),
\label{Vwdotineq3}\\
& \text{where}~  \varrho_v \triangleq  \frac{\min(\Lambda_v, \alpha_{v0}, \alpha_{v1}, \alpha_{w2})}{\max(m/2, 1/2)} >0 \nonumber  \\
  & \qquad \quad \varrho_w \triangleq  \frac{\min(\Lambda_w, \alpha_{w0}, \alpha_{w1}, \alpha_{v2})}{\max(J/2, 1/2)} > 0 \nonumber.
\end{align}
Combining (\ref{Vvdotineq3}) and (\ref{Vwdotineq3}), the time derivative of the overall Lyapunov function $\dot{V}$ can be obtained as
\begin{align}
    \dot{V} =& -\varrho V + (1/2)(\alpha_{v0} {{K}^*_{v0}}^2  + \alpha_{v1} {{K}^*_{v1}}^2 + \alpha_{w2} {{K}^*_{w2}}^2) \nonumber\\
    & + (1/2) (\alpha_{w0} {{K}^*_{w0}}^2  + \alpha_{w1} {{K}^*_{w1}}^2 + \alpha_{v2} {{K}^*_{v2}}^2), \label{new3}
\end{align}
where $\varrho \triangleq \min(\varrho_v, \varrho_w)$. Defining a scalar $\kappa$ such that $0<\kappa<\varrho$, $\dot{V}$ in (\ref{new3}) is further simplified to
\begin{align}
    \dot{V} \leq & -\kappa V -(\varrho - \kappa)V + \frac{1}{2}(\alpha_{v0} {{K}^*_{v0}}^2  + \alpha_{v1} {{K}^*_{v1}}^2 + \alpha_{w2} {{K}^*_{w2}}^2) \nonumber\\
    & + (1/2)(\alpha_{w0} {{K}^*_{w0}}^2  + \alpha_{w1} {{K}^*_{w1}}^2 + \alpha_{v2} {{K}^*_{v2}}^2)).
\end{align}
Defining a scalar $\beta \triangleq \frac{\sum_{i=0}^{2} (\alpha_{vi} {{K}^*_{vi}}^2 + \alpha_{wi} {{K}^*_{wi}}^2) }{2(\varrho - \kappa)} $, it can been noted that $\dot{V}(t) < -\kappa V(t)$ when $V(t)\geq \beta$, leading to
 \begin{equation}
     V \leq \max ( V(0), \beta ),~ \forall t>0,
 \end{equation}
and hence, the closed-loop system remains UUB.

To avoid chattering due to discontinuity in control law, the ‘$sgn$’ functions in (\ref{force_law}) and (\ref{torque_law}) are often replaced by a ‘saturation’/sigmoid functions which leads to minor modifications in the stability analysis without altering the overall UUB result and hence omitted to avoid repetition (cf. \cite{roy2019overcoming, asif2014adaptive}).

\section{Simulation Results}
\subsection{Simulation Scenario}
\begin{figure}[h]
\begin{center}
    \includegraphics[scale=0.50]{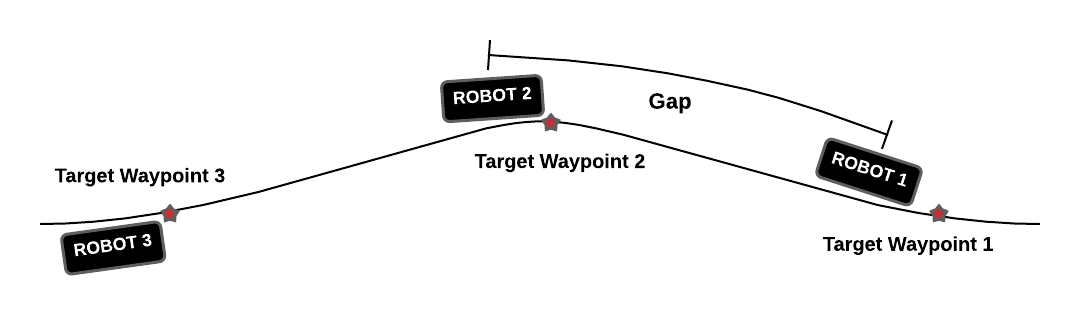}
    \newline
    \caption{\small Multi-Robot Platoon Representation}
    \label{formation}
\end{center}
\end{figure}

The multi-robot platoon system operates in a decentralized manner, where each robot only requires the state information of its predecessor robot to maintain a formation (cf. Fig. \ref{formation}). The desired waypoints for the platoon are generated using Algorithm \ref{alg:algo1}, which stores the waypoints in an array. Each robot calculates its desired state based on the index number of its local leader robot in the array. Instead of calculating the shortest distance between consecutive robots, the gap between them is calculated along the path. This approach ensures that the platoon maintains a consistent formation while navigating along a given path. By using this decentralized formation control system, the multi-robot platoon can efficiently navigate through challenging environments and achieve their desired goals. The waypoints generated by Algorithm \ref{alg:algo1} provide a clear and precise path for the robots to follow, enabling them to maintain a consistent formation throughout their mission.

\begin{algorithm}[]
\caption{Target way-point calculation}
\label{alg:algo1}
\textbf{Initialisation;}\\
initialize $d$ = 0\\
initialize $gap_{des}$\\
$cx$ = array containing x-coordinates of  trajectory\\
$cy$ = array containing y-coordinates of trajectory\\
$i  \gets  index_{leader} $\\

\While {$d < gap_{des}$}{
    $d += [(cx[i]-cx[i-1])^2 + (cy[i]-cy[i-1])^2]^{0.5}$\\
    $i = i - 1$
}
\textbf{return} follower $x_r = cx[i], y_r = cy[i]$
\end{algorithm}

\begin{figure}[]
	\centering
    \includegraphics[scale = 0.3]{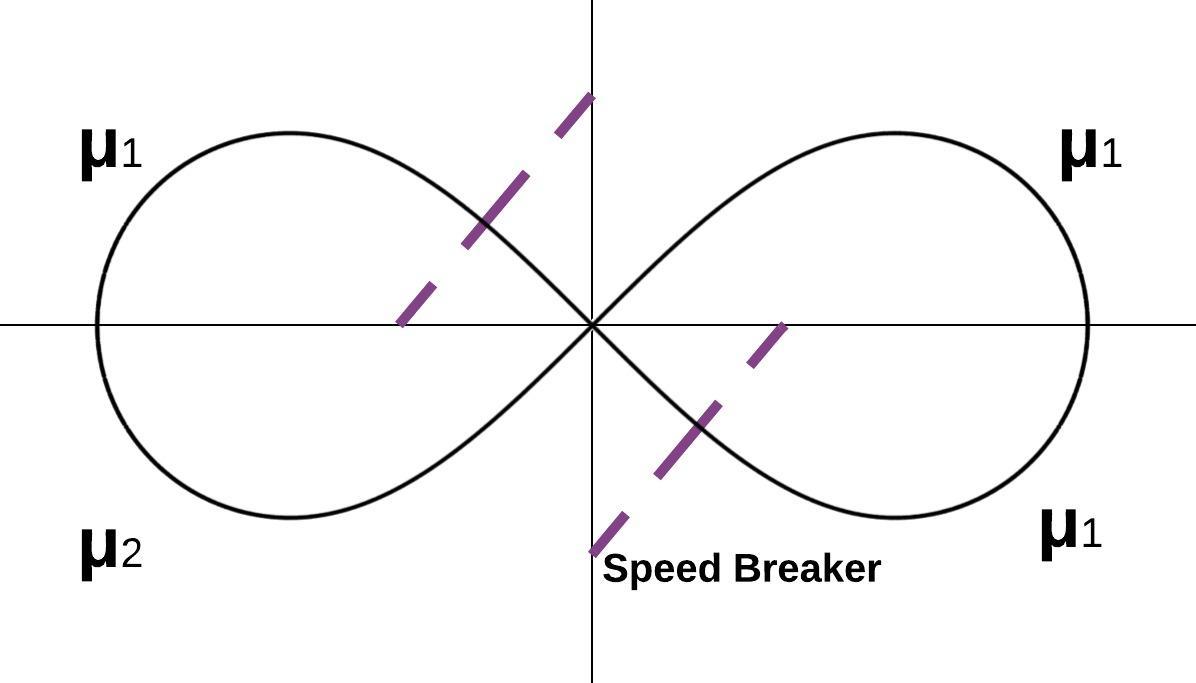} 
    \caption{\small Custom Arena} \normalsize
    \label{arena}
\end{figure}

The performance of the proposed adaptive sliding mode controller
(ASMC) is compared with standard ASMC \cite{asif2014adaptive} on a Gazebo simulation platform using open-source TurtleBot3 robot model \cite{turtlebot} and Teeterbot plugin \cite{teeterbot} to give a torque command to each motor using  (\ref{wheeltorque}).
To evaluate the performance of the controller, a custom arena has been created in Gazebo as world which is divided into four quadrants (cf. Fig. \ref{arena}): surface longitudinal friction $\mu_1 = 0.1$ is selected for 1$^{st}$, 2$^{nd}$ and 4$^{th}$ quadrant, while it is kept as $\mu_2$ = 0.13 for the 3$^{rd}$ quadrant; while lateral friction is kept as $0.1$ in all quadrants to avoid lateral slippage of robots. In addition, two speed breakers have been placed to give sudden interrupt. 

Three robots (Robot 1, 2 and 3) are used for the formation control, where they are required to follow a figure-of-eight like path as in Fig. \ref{arena}. The lead robot is initially placed at the co-ordinate (14, 0), while follower robots are initially kept at the desired distance of $1$m between them. 

\subsection{Parameter Selection}
For simulation, the kinematic control parameters are selected as: $k_1$ = 5, $k_2$ = 3, $k_3$ = 2,
$v_d = 2$ m/sec for all robots. The control parameters of the proposed ASMC are selected to be: $\phi_v = 0.5, \phi_w = 0.1, \Lambda_v = 3, \Lambda_w = 2, \hat{K}_{v0}(0) =  \hat{K}_{v1}(0) = \hat{K}_{w2}(0) =  \hat{K}_{w0}(0) = \hat{K}_{w1}(0) = \hat{K}_{v2}(0) = 0.01, \alpha_{v0} = 2.5, \alpha_{v1} = 2.5,\alpha_{w2} = 3,\alpha_{w0} = 5,\alpha_{w1} = 5,\alpha_{v2} = 1.5$ for all robots. For parity, similar sliding variable and similar kinematic control parameters are designed for the standard ASMC \cite{asif2014adaptive}. 

\subsection{Results and Analysis}

\begin{figure}[]
	\centering
    \includegraphics[scale = 0.27]{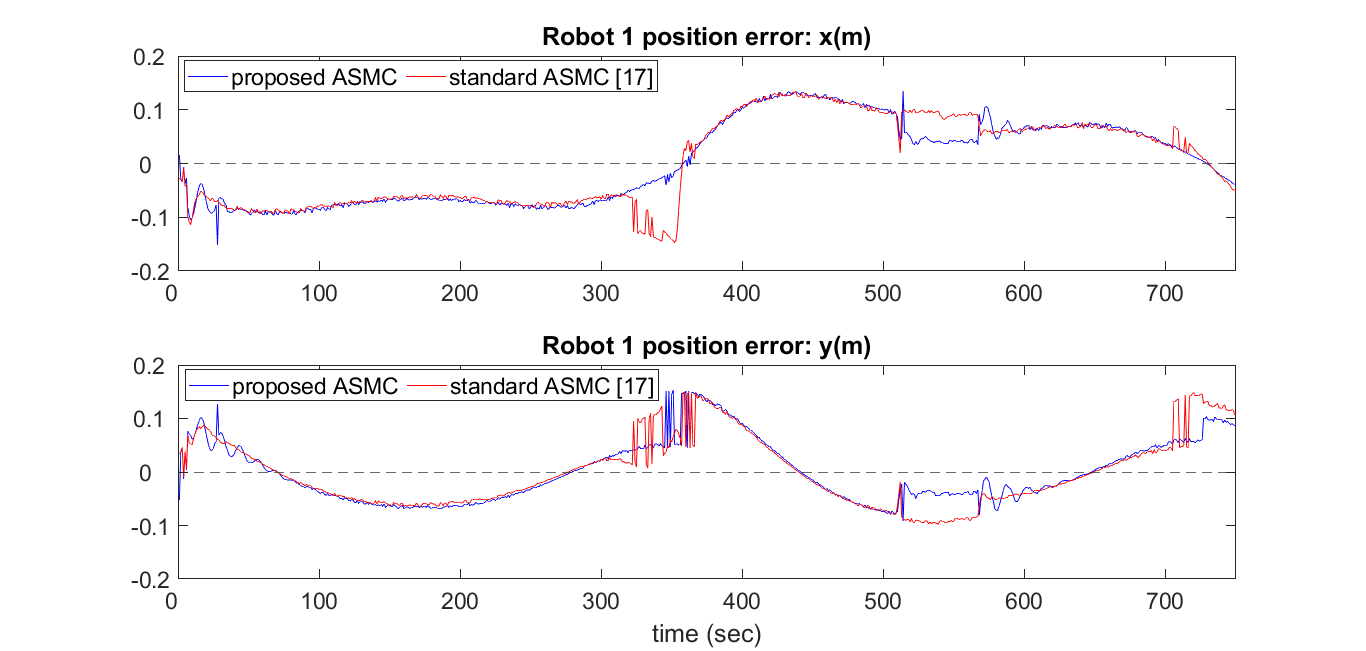} 
    \caption{\small Position tracking error comparison for Robot 1.} \normalsize
    \label{dx1dy1}
\end{figure}

\begin{figure}[]
	\centering
    \includegraphics[scale = 0.27]{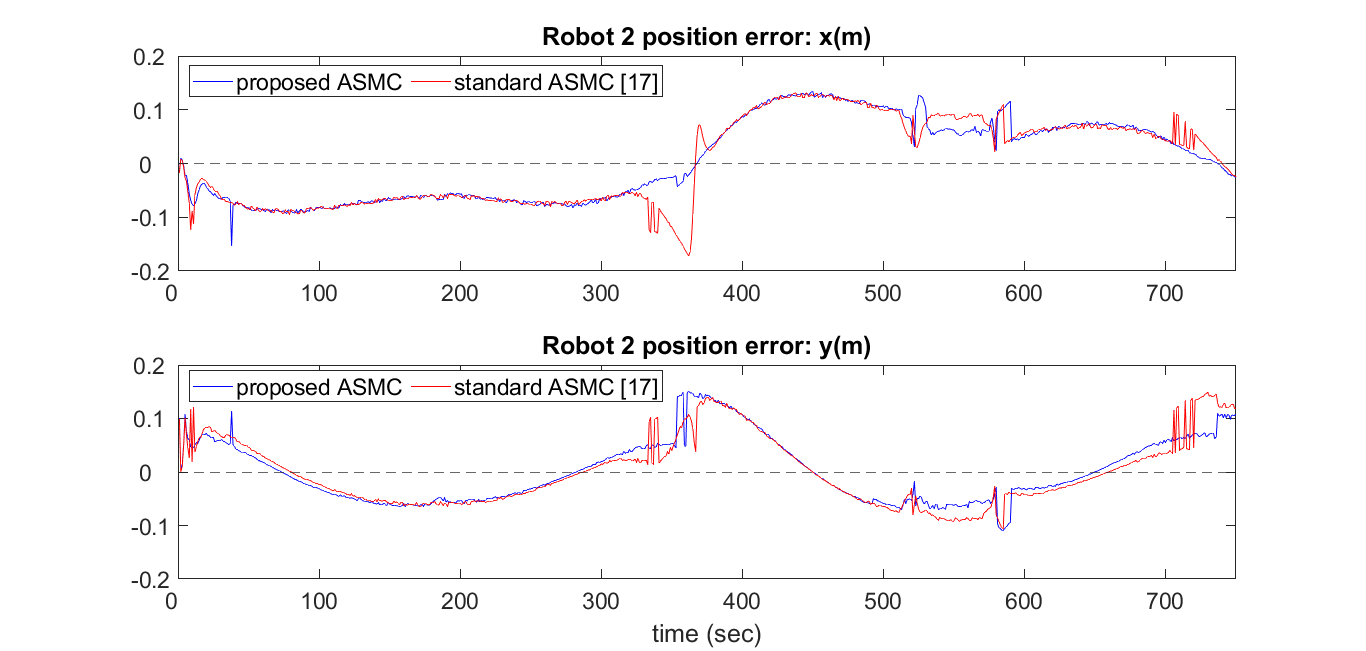} 
    \caption{\small Position tracking error comparison for Robot 2.} \normalsize
    \label{dx2dy2}
\end{figure}

\begin{figure}[]
	\centering
    \includegraphics[scale = 0.27]{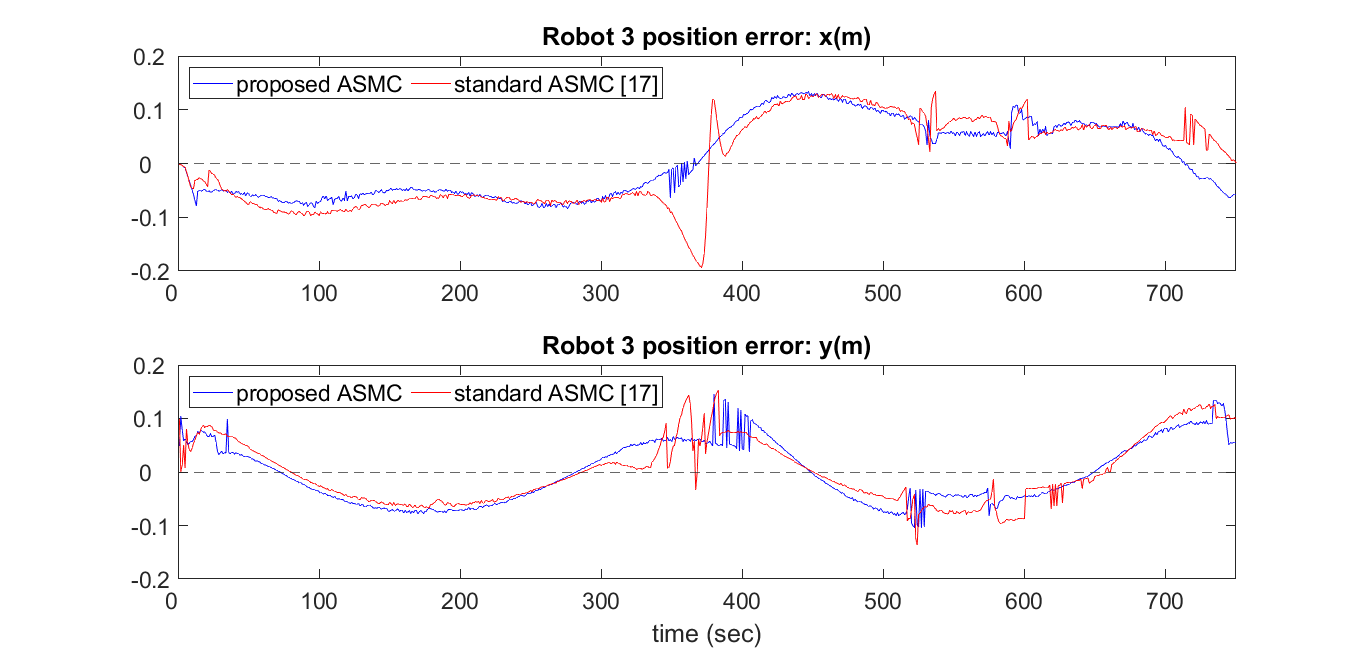} 
    \caption{\small Position tracking error comparison for Robot 3.} \normalsize
    \label{dx3dy3}
\end{figure}

\begin{figure}[]
	\centering
    \includegraphics[scale = 0.26]{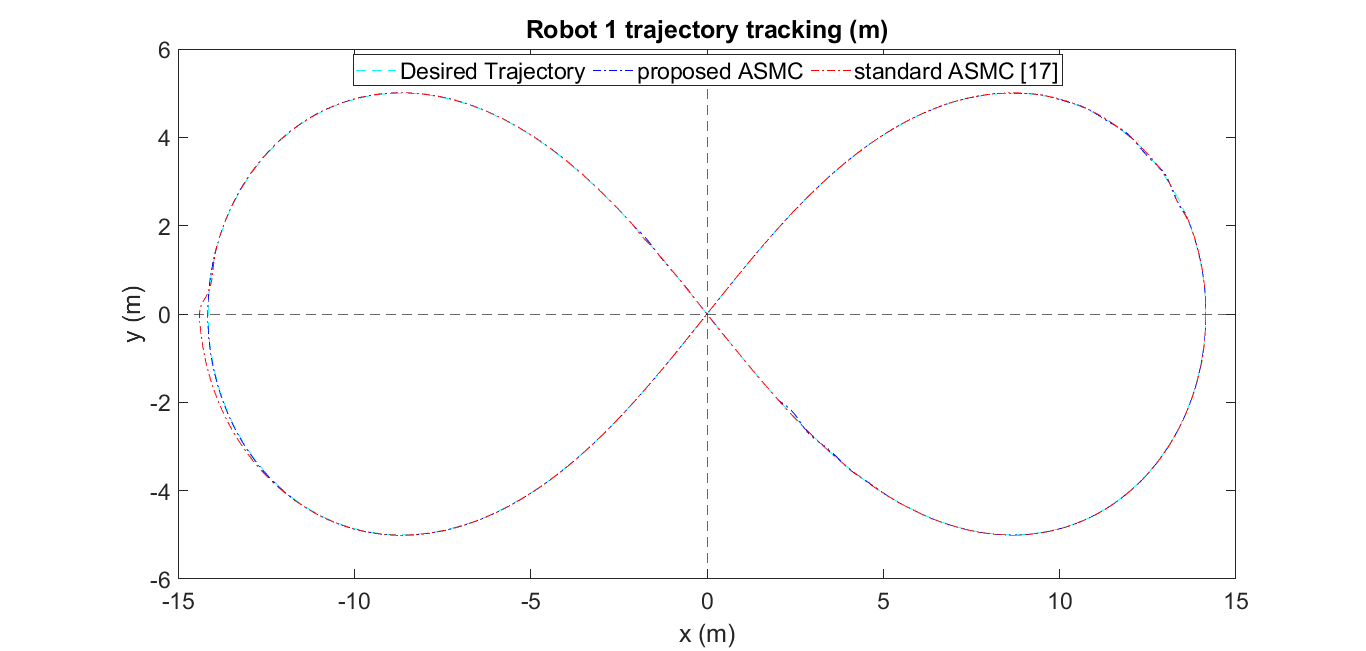} 
    \caption{\small Trajectory tracking comparison for Robot 1.} \normalsize
    \label{robot1}
\end{figure}

\begin{figure}[]
	\centering
    \includegraphics[scale = 0.26]{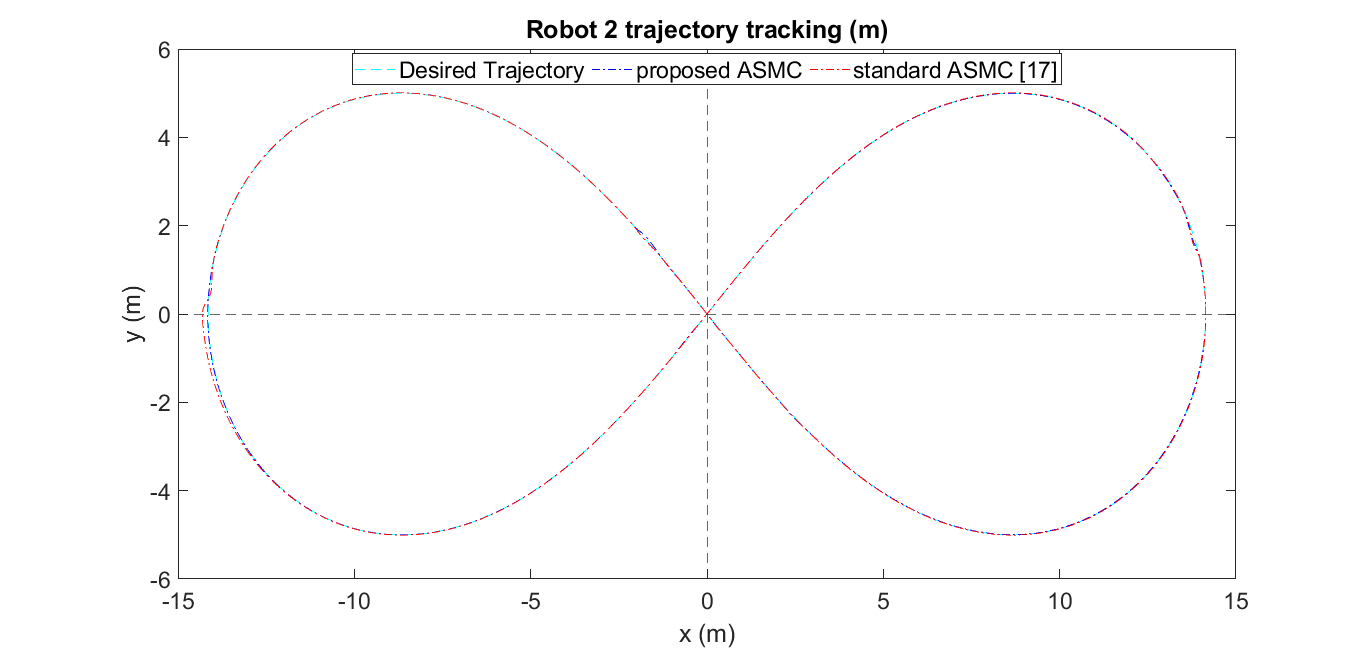} 
    \caption{\small Trajectory tracking comparison for Robot 2.} \normalsize
    \label{robot2}
\end{figure}

\begin{figure}[]
	\centering
    \includegraphics[scale = 0.26]{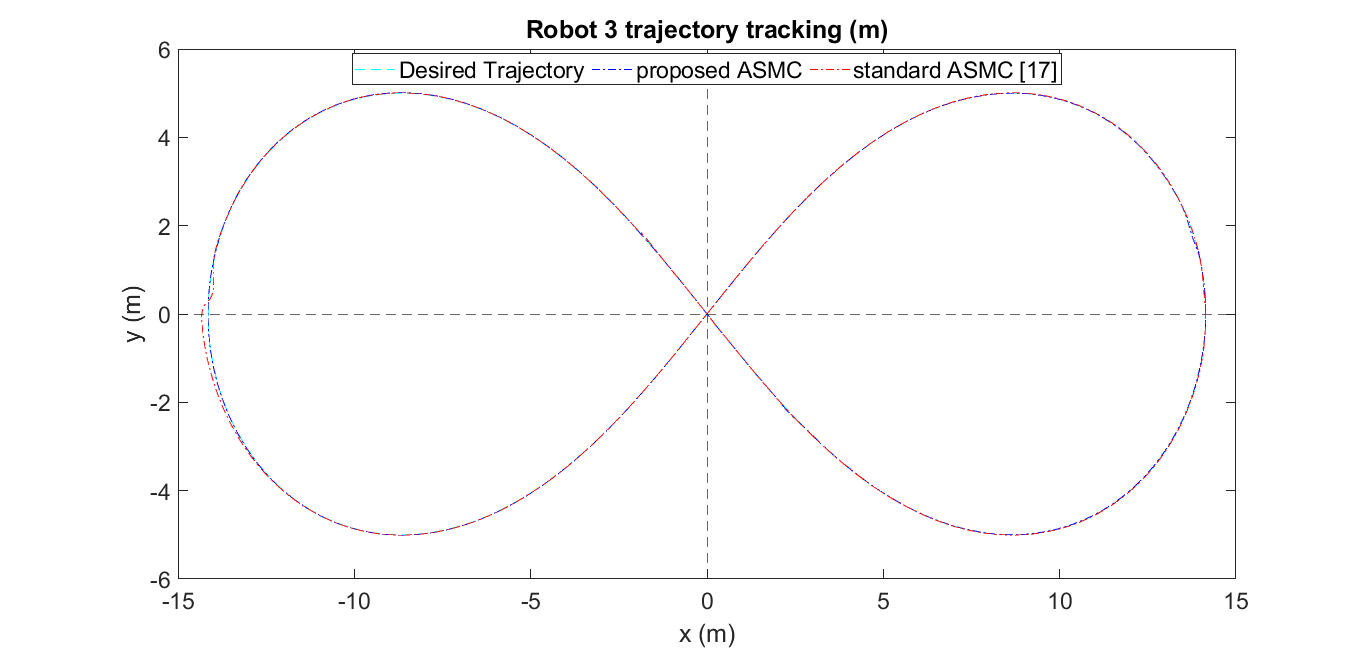} 
    \caption{\small Trajectory tracking comparison for Robot 3.} \normalsize
    \label{robot3}
\end{figure}

The study compared the performance of two controllers, the proposed Adaptive Sliding Mode Controller (ASMC) and the standard ASMC \cite{asif2014adaptive} via Figs. \ref{dx1dy1}, \ref{dx2dy2} and \ref{dx3dy3}, in terms of position tracking error for three robots, labeled 1, 2, and 3. The analysis was carried out in four quadrants. In the third quadrant, between $300<t<400$, all robots experienced a sharp turn and high friction surface, causing an increase in tracking error. The results show that the standard ASMC \cite{asif2014adaptive} performed worse than the proposed ASMC during this period, as it was not designed to handle changes in state-dependent friction components. A similar trend was observed between $500<t<600$, when the robots encountered two speed breakers. The proposed controller showed fewer spikes in the tracking error profile during this period. The performance of both controllers was found to be similar in the first, second, and fourth quadrants. This was confirmed by path tracking performances shown in Figures  \ref{robot1}, \ref{robot2} and \ref{robot3}. Overall, the results indicate that the proposed controller performs better in the third quadrant, where changes in friction components are significant, while the standard and proposed controllers have similar performance in the other quadrants.

\begin{figure}[]
	\centering
    \includegraphics[scale = 0.26]{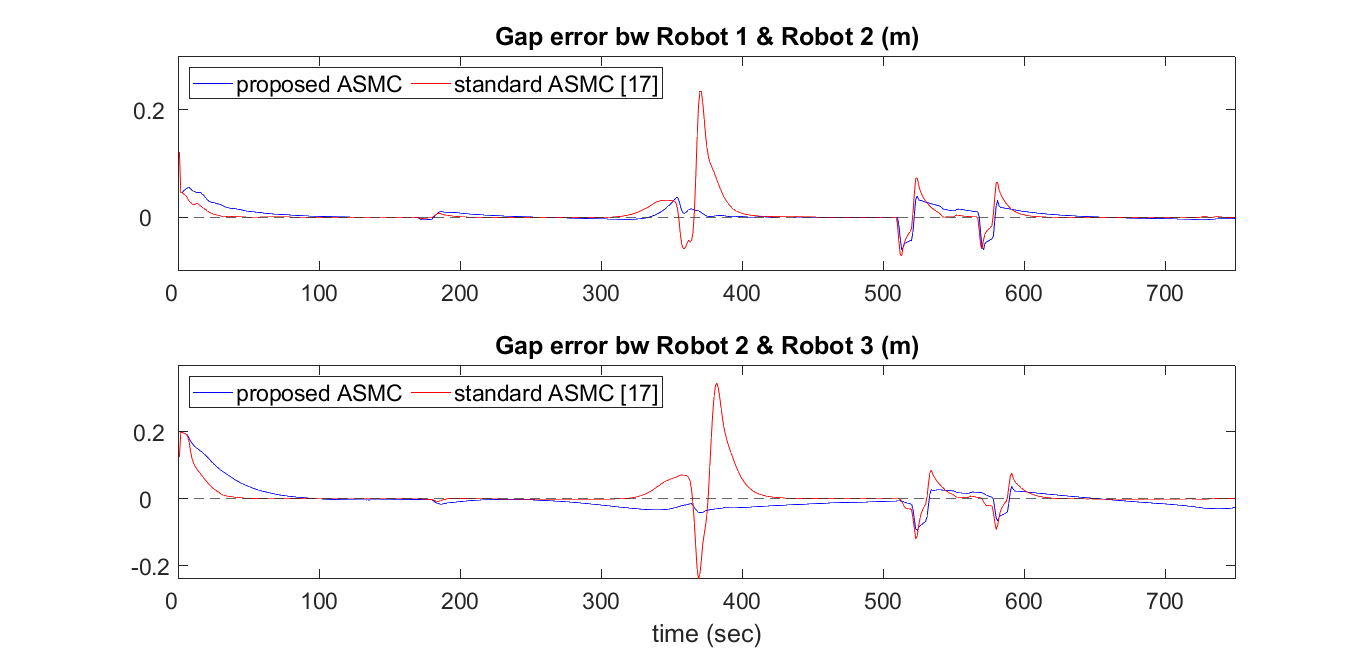} 
    \caption{\small (a) Gap error comparison  between Robot 1 and Robot 2. (b) Gap error comparison between Robot 2 and Robot 3.} \normalsize
    \label{gaperror}
\end{figure}

Additionally, in Figure \ref{gaperror}, the difference between the intended gap and the actual gap of two consecutive robots, namely Robot 1-Robot 2 and Robot 2-Robot 3, is illustrated, which is called the gap error. At around $t = 350$ seconds, it can be observed that when the robots are in the third quadrant and encounter a high variation in friction, the standard ASMC \cite{asif2014adaptive} exhibits a significantly greater gap error than the proposed ASMC. To provide more conclusive evidence, Tables \ref{PostionError} and \ref{GapEerror} present the performance of both controllers in terms of root mean-squared (RMS) error. The data in these tables indicate that the proposed ASMC achieves superior tracking accuracy while preserving the desired distance between the robots.

\begin{table}[]
\caption{Performance Comparison for Trajectory Tracking}
\centering  
\begin{tabular}{l c  rr}  
\hline\hline                       
\\ [-2ex]
No. & Position & standard ASMC [17]  & proposed ASMC  \\  
& & RMS error (m) & RMS error(m) \\ [0.5ex]
\hline
\\ [-1ex]

&x & $0.081$ & $0.076$  \\ [-1ex]
\raisebox{1.5ex}{Robot1} & y &  $0.063$ & $0.056$ \\[1.5ex]

&x & $0.079$ & $0.076$  \\ [-1ex]
\raisebox{1.5ex}{Robot2} & y &  $0.062$ & $0.058$ \\[1.5ex]

&x & $0.080$ & $0.070$  \\ [-1ex]
\raisebox{1.5ex}{Robot3} & y&  $0.060$ & $0.056$ \\[1.5ex]

\hline                          
\end{tabular}
\label{PostionError}
\end{table}


\begin{table}[h]
\caption{Performance Comparison for Gap Maintenance}
\centering  
\begin{tabular}{l c  rr}  
\hline\hline                     
\\ [-2ex]
Gap Error b/w Robots & standard ASMC [17] & proposed ASMC \\  
&  RMS error (m) & RMS error(m) \\ [0.5ex]
\hline
\\ [-1ex]

Robot 1 and  Robot 2 & $0.026$ & $0.014$  \\[0.5ex]
Robot 2 and Robot 3 &  $0.050$ & $0.036$ \\ [0.5ex]

\hline                          
\end{tabular}
\label{GapEerror}
\end{table}

\section{Conclusions}
A new method has been suggested for controlling the formation of a platoon of autonomous wheeled mobile robots, which can manage factors such as external interferences, uncertainties in parameters, and variations in friction between the tire and the surface without any prior knowledge of them. The proposed adaptive sliding mode controller uses Lyapunov function to establish stability of the closed-loop system, with the help of Uniformly Ultimately Bounded concept. The effectiveness of the controller was tested under different conditions using Gazebo simulation, and the results showed significant enhancements in the performance of the platoon, in terms of both trajectory tracking and maintaining a fixed safe distance between the robots, when compared to existing methods.

The controller's ability to adjust to unknown factors, such as external disturbances, frictional differences, and parametric uncertainties, is a key feature that sets it apart from other techniques. By using a sliding mode approach, the controller's design ensures robustness and adaptation to such variations. Additionally, the Lyapunov function used in the analysis guarantees that the closed-loop system is stable, and the Uniformly Ultimately Bounded notion ensures that the solutions are bounded and convergent. The effectiveness of the proposed method was demonstrated through simulations, where it was observed that the proposed controller performed better than the existing methods, achieving high accuracy in trajectory tracking and maintaining a safe distance between the robots.

Future work involve implementing the proposed adaptive sliding mode controller on a real-world platform and conducting experiments to evaluate its performance with different slope and payloads. The controller could also be further optimized for improved energy efficiency and faster response times. Additionally, the controller's scalability could be investigated to determine its effectiveness in controlling larger platoons of mobile robots. Finally, the proposed method could be extended to other types of autonomous systems, such as aerial drones or underwater vehicles.

\section*{Related Publications}

\begin{enumerate}
    \item \textbf{Adaptive Sliding Mode Control for Autonomous Vehicle Platoon under Unknown Friction Forces} \cite{yadav2021adaptive} \\
    \textbf{R. D. Yadav}*, V. Sankaranarayanan and S. Roy,
    \\
    \textit{IEEE 20th International Conference on advanced robotics (ICAR 2021)}
\end{enumerate}
\begingroup
\let\clearpage\relax
\section*{Other Publications}
\endgroup

\begin{enumerate}
    \item \textbf{Efficient Manoeuvring of Quadrotor under Constrained Space and Predefined Accuracy} \cite{ganguly2021efficient} \\
     S. Ganguly, V. Sankaranarayanan, B. V. S. G. Suraj, \textbf{R. D. Yadav} and S. Roy \\
    \textit{ The 2021 IEEE/RSJ International Conference on Intelligent Robots
and Systems (IROS 2021)}

    \item \textbf{Robust Manoeuvring of Quadrotor under Full State Constraints } \cite{ganguly2022robust}\\
    S. Ganguly, V. Sankaranarayanan, B. V. S. G. Suraj, \textbf{R. D. Yadav} and S. Roy  \\
    \textit{Automatic Control and Dynamical Optimization Society (ACDOS 2022) }
    
    \item \textbf{Robustifying Payload Carrying Operations for Quadrotors Under Time-Varying State Constraints and Uncertainty} \cite{sankaranarayanan2022robustifying}\\
     V. Shankaranarayanan, \textbf{Rishabh Dev Yadav}, R. K. Swyampakula, S. Ganguly and S. Roy\\
    \textit{ IEEE Robotics and Automation Letters, vol. 7, no. 2, pp. 4885-4892. (RAL)}
    
    \item \textbf{CCO-VOXEL: Chance
    Constrained Optimization over Uncertain Voxel-Grid Representation for Safe Trajectory Planning} \cite{harithas2022cco} \\
    Sudarshan S Harithas, \textbf{R. D. Yadav}, Deepak Singh, Arun Kumar Singh, K Madhava Krishna
    \\
    \textit{ IEEE International
    Conference on Robotics and Automation (ICRA 2022)}

    \item \textbf{Introducing Scissor Mechanism based Novel Reconfigurable Quadrotor: Design, Modelling and Control} \cite{suraj2022introducing}\\
    B. V. S. G Suraj, Viswa N. Sankaranarayanan, \textbf{Rishabh Dev Yadav} and Spandan Roy, \\
    \textit{IEEE International Conference on Robotics and Biomimetics  (ROBIO 2022)}

    \item \textbf{Adaptive Artificial Time Delay Control for Quadrotors under State-dependent Unknown Dynamic} \cite{dantu2022adaptive}\\
    Swati Dantu, \textbf{R. D. Yadav}, Spandan Roy, Jinoh Lee and Simone Baldi \\
    \textit{IEEE International Conference on Robotics and Biomimetics  (ROBIO 2022)}

    \item \textbf{Adaptive Anti-swing Control for Clasping Operations in Quadrotors with Cable-suspended Payload} \cite{dantu2023adaptive}\\
    S. Dantu, \textbf{R. D. Yadav}, A. Rachakonda, S. Roy, S. Baldi \\
    \textit{IEEE Conference on Decision and Control (CDC 2023)}

    \item \textbf{Adaptive Control of Quadrotor under Actuator Loss and Unknown State-dependent Dynamics} \cite{gupta2024adaptive}\\
    Saksham Gupta, Amitabh Sharma, Aditya Mulgundkar, \textbf{Rishabh Dev Yadav}, Spandan Roy \\
    \textit{IEEE International Conference on Automation Science and Engineering (CASE)}

    \item \textbf{Modular adaptive aerial manipulation under unknown dynamic coupling forces} \cite{yadav2024modular}\\
    \textbf{Rishabh Dev Yadav}, Swati Dantu, Wei Pan, Sihao Sun, Spandan Roy, Simone Baldi \\
    \textit{IEEE/ASME Transactions on Mechatronics}

    \item \textbf{Adaptive Tracking and Anti-Swing Control of Quadrotors Carrying Suspended Payload Under State-Dependent Uncertainty} \cite{dantu2024adaptive}\\
    Swati Dantu, \textbf{Rishabh Dev Yadav}, Ananth Rachakonda, Spandan Roy, Simone Baldi \\
    \textit{IEEE/ASME Transactions on Mechatronics}

    \item \textbf{An Integrated Approach to Aerial Grasping: Combining a Bistable Gripper with Adaptive Control} \cite{yadav2025integrated}\\
    \textbf{Rishabh Dev Yadav}, Brycen Jones, Saksham Gupta, Amitabh Sharma, Jiefeng Sun, Jianguo Zhao, Spandan Roy \\
    \textit{IEEE/ASME Transactions on Mechatronics}

    \item \textbf{Impedance and Stability Targeted Adaptation for Aerial Manipulator with Unknown Coupling Dynamics} \cite{sharma2025impedance}\\
    Amitabh Sharma, Saksham Gupta, Shivansh Pratap Singh, \textbf{Rishabh Dev Yadav}, Hongyu Song, Wei Pan, Spandan Roy, Simone Baldi \\
    \textit{IEEE  International Conference on Control, Automation and Systems (ICCAS)}
    
\end{enumerate}
\bibliographystyle{IEEEtran}
\bibliography{references} 

@article{shukla2021robust,
  title={Robust adaptive control of steer-by-wire systems under unknown state-dependent uncertainties},
  author={Shukla, Harsh and Roy, Spandan and Gupta, Satyam},
  journal={International Journal of Adaptive Control and Signal Processing},
  year={2021},
  publisher={Wiley Online Library}
}

@article{roy2021adaptive,
  title={An Adaptive Control Framework for Underactuated Switched Euler-Lagrange Systems},
  author={Roy, Spandan and Baldi, Simone and Ioannou, Petros A},
  journal={IEEE Transactions on Automatic Control},
  year={2021},
  publisher={IEEE}
}

@article{roy2020adaptive,
  title={On adaptive sliding mode control without a priori bounded uncertainty},
  author={Roy, Spandan and Baldi, Simone and Fridman, Leonid M},
  journal={Automatica},
  volume={111},
  pages={108650},
  year={2020},
  publisher={Elsevier}
}

@article{roy2020towards,
  title={Towards structure-independent stabilization for uncertain underactuated Euler--Lagrange systems},
  author={Roy, Spandan and Baldi, Simone},
  journal={Automatica},
  volume={113},
  pages={108775},
  year={2020},
  publisher={Elsevier}
}

@article{roy2019overcoming,
  title={Overcoming the underestimation and overestimation problems in adaptive sliding mode control},
  author={Roy, Spandan and Roy, Sayan Basu and Lee, Jinoh and Baldi, Simone},
  journal={IEEE/ASME Transactions on Mechatronics},
  volume={24},
  number={5},
  pages={2031--2039},
  year={2019},
  publisher={IEEE}
}

@article{utkin2013adaptive,
  title={Adaptive sliding mode control with application to super-twist algorithm: Equivalent control method},
  author={Utkin, Vadim I and Poznyak, Alex S},
  journal={Automatica},
  volume={49},
  number={1},
  pages={39--47},
  year={2013},
  publisher={Elsevier}
}

@inproceedings{darmanin2017review,
  title={A review on multi-robot systems categorised by application domain},
  author={Darmanin, Rachael N and Bugeja, Marvin K},
  booktitle={25th Mediterranean Conference on Control and Automation},
  pages={701--706},
  year={2017},
  organization={IEEE}
}

@article{arai2002advances,
  title={Advances in multi-robot systems},
  author={Arai, Tamio and Pagello, Enrico and Parker, Lynne E and others},
  journal={IEEE Transactions on Robotics and Automation},
  volume={18},
  number={5},
  pages={655--661},
  year={2002},
  publisher={Citeseer}
}

@incollection{ismail2018survey,
  title={A survey and analysis of cooperative multi-agent robot systems: Challenges and directions},
  author={Ismail, Zool Hilmi and Sariff, Nohaidda},
  booktitle={Applications of Mobile Robots},
  pages={8--14},
  year={2018},
  publisher={IntechOpen}
}

@article{cortes2017coordinated,
  title={Coordinated control of multi-robot systems: A survey},
  author={Cort{\'e}s, Jorge and Egerstedt, Magnus},
  journal={SICE Journal of Control, Measurement, and System Integration},
  volume={10},
  number={6},
  pages={495--503},
  year={2017},
  publisher={The Society of Instrument and Control Engineers}
}

@inproceedings{cai2012survey,
  title={A survey on multi-robot systems},
  author={Cai, Yifan and Yang, Simon X},
  booktitle={World Automation Congress 2012},
  pages={1--6},
  year={2012},
  organization={IEEE}
}

@inproceedings{vcerkala2014mobile,
  title={Mobile Robot Dynamics with Friction in Simulink},
  author={{\v{C}}erkala, J and Jadlovska, A},
  booktitle={Proceedings of the 22th Annual Conference Proceedings of the International Scientific Conference—Technical Computing Bratislava},
  pages={1--10},
  year={2014}
}

@article{soni2018formation,
  title={Formation control for a fleet of autonomous ground vehicles: A survey},
  author={Soni, Aakash and Hu, Huosheng},
  journal={Robotics},
  volume={7},
  number={4},
  pages={67},
  year={2018},
  publisher={Multidisciplinary Digital Publishing Institute}
}

@article{iurian2005identification,
  title={Identification of a system with dry friction},
  author={Iurian, Claudiu and Ikhouane, Fay{\c{c}}al and Rodellar Bened{\'e}, Jos{\'e} and Gri{\~n}{\'o} Cubero, Robert},
  year={2005}
}

@inproceedings{berntorp2019steering,
  title={Steering of autonomous vehicles based on friction-adaptive nonlinear model-predictive control},
  author={Berntorp, Karl and Quirynen, Rien and Di Cairano, Stefano},
  booktitle={American Control Conference},
  pages={965--970},
  year={2019},
  organization={IEEE}
}

@article{antonelli2006kinematic,
  title={Kinematic control of platoons of autonomous vehicles},
  author={Antonelli, Gianluca and Chiaverini, Stefano},
  journal={IEEE Transactions on Robotics},
  volume={22},
  number={6},
  pages={1285--1292},
  year={2006},
  publisher={IEEE}
}

@article{zhang2020analysis,
  title={Analysis and design on intervehicle distance control of autonomous vehicle platoons},
  author={Zhang, Jilie and Feng, Tao and Yan, Fei and Qiao, Shaojie and Wang, Xiaomin},
  journal={ISA Transactions},
  volume={100},
  pages={446--453},
  year={2020},
  publisher={Elsevier}
}

@article{zhao2018robust,
  title={Robust approximate constraint-following control for autonomous vehicle platoon systems},
  author={Zhao, Xiaomin and Chen, YH and Zhao, Han},
  journal={Asian Journal of Control},
  volume={20},
  number={4},
  pages={1611--1623},
  year={2018},
  publisher={Wiley Online Library}
}

@article{tuchner2017vehicle,
  title={Vehicle platoon formation using interpolating control: A laboratory experimental analysis},
  author={Tuchner, Alon and Haddad, Jack},
  journal={Transportation Research Part C: Emerging Technologies},
  volume={84},
  pages={21--47},
  year={2017},
  publisher={Elsevier}
}

@article{xu2019distributed,
  title={Distributed formation control of homogeneous vehicle platoon considering vehicle dynamics},
  author={Xu, Liwei and Zhuang, Weichao and Yin, Guodong and Bian, Chentong},
  journal={International Journal of Automotive Technology},
  volume={20},
  number={6},
  pages={1103--1112},
  year={2019},
  publisher={Springer}
}

@article{dai2017platoon,
  title={Platoon formation control with prescribed performance guarantees for USVs},
  author={Dai, Shi-Lu and He, Shude and Lin, Hai and Wang, Cong},
  journal={IEEE Transactions on Industrial Electronics},
  volume={65},
  number={5},
  pages={4237--4246},
  year={2017},
  publisher={IEEE}
}

@article{chen2009design,
  title={Design and implementation of an adaptive sliding-mode dynamic controller for wheeled mobile robots},
  author={Chen, Chih-Yang and Li, Tzuu-Hseng S and Yeh, Ying-Chieh and Chang, Cha-Cheng},
  journal={Mechatronics},
  volume={19},
  number={2},
  pages={156--166},
  year={2009},
  publisher={Elsevier}
}

@article{asif2014adaptive,
  title={Adaptive sliding mode dynamic controller with integrator in the loop for nonholonomic wheeled mobile robot trajectory tracking},
  author={Asif, Muhammad and Khan, Muhammad Junaid and Cai, Ning},
  journal={International Journal of Control},
  volume={87},
  number={5},
  pages={964--975},
  year={2014},
  publisher={Taylor \& Francis}
}

@article{koubaa2018adaptive,
  title={Adaptive sliding mode control for trajectory tracking of nonholonomic mobile robot with uncertain kinematics and dynamics},
  author={Koubaa, Yasmine and Boukattaya, Mohamed and Damak, Tarak},
  journal={Applied Artificial Intelligence},
  volume={32},
  number={9-10},
  pages={924--938},
  year={2018},
  publisher={Taylor \& Francis}
}

@article{fierro1997control,
  title={Control of a nonholomic mobile robot: Backstepping kinematics into dynamics},
  author={Fierro, Rafael and Lewis, Frank L},
  journal={Journal of robotic systems},
  volume={14},
  number={3},
  pages={149--163},
  year={1997},
  publisher={Wiley Online Library}
}

@article{garcia2017tracking,
  title={Tracking control for mobile robots considering the dynamics of all their subsystems: Experimental implementation},
  author={Garc{\'\i}a-S{\'a}nchez et al., Jos{\'e} Rafael},
  journal={Complexity},
  volume={2017},
  year={2017},
  publisher={Hindawi}
}

@article{liu2020adaptive,
  title={Adaptive sliding mode based disturbance attenuation tracking control for wheeled mobile robots},
  author={Liu, Kang and Gao, Hongbo and Ji, Haibo and Hao, Zhengyuan},
  journal={International Journal of Control, Automation and Systems},
  volume={18},
  number={5},
  pages={1288--1298},
  year={2020},
  publisher={Springer}
}

@misc{teeterbot,
    author    = "Robustify",
    title     = "TeeterBot: Self-Balancing Robot",
    url       = "https://github.com/robustify/teeterbot"
}

@misc{turtlebot,
    author    = "ROBOTIS-GIT",
    title     = "TurtleBot3",
    url       = "https://github.com/ROBOTIS-GIT/turtlebot3"
}

@inproceedings{yadav2021adaptive,
  title={Adaptive sliding mode control for autonomous vehicle platoon under unknown friction forces},
  author={Yadav, Rishabh Dev and Sankaranarayanan, Viswa N and Roy, Spandan},
  booktitle={2021 20th International Conference on Advanced Robotics (ICAR)},
  pages={879--884},
  year={2021},
  organization={IEEE}
}

@article{sankaranarayanan2022robustifying,
  title={Robustifying payload carrying operations for quadrotors under time-varying state constraints and uncertainty},
  author={Sankaranarayanan, Viswa N and Yadav, Rishabh D and Swayampakula, Rahul K and Ganguly, Sourish and Roy, Spandan},
  journal={IEEE Robotics and Automation Letters},
  volume={7},
  number={2},
  pages={4885--4892},
  year={2022},
  publisher={IEEE}
}

@inproceedings{harithas2022cco,
  title={Cco-voxel: Chance constrained optimization over uncertain voxel-grid representation for safe trajectory planning},
  author={Harithas, Sudarshan S and Yadav, Rishabh Dev and Singh, Deepak and Singh, Arun Kumar and Krishna, K Madhava},
  booktitle={2022 International Conference on Robotics and Automation (ICRA)},
  pages={11087--11093},
  year={2022},
  organization={IEEE}
}

@inproceedings{suraj2022introducing,
  title={Introducing scissor mechanism based novel reconfigurable quadrotor: Design, modelling and control},
  author={Suraj, BVSG and Sankaranarayanan, Viswa N and Yadav, Rishabh Dev and Roy, Spandan},
  booktitle={2022 IEEE International Conference on Robotics and Biomimetics (ROBIO)},
  pages={2231--2236},
  year={2022},
  organization={IEEE}
}

@inproceedings{dantu2022adaptive,
  title={Adaptive artificial time delay control for quadrotors under state-dependent unknown dynamics},
  author={Dantu, Swati and Yadav, Rishabh Dev and Roy, Spandan and Lee, Jinoh and Baldi, Simone},
  booktitle={2022 IEEE International Conference on Robotics and Biomimetics (ROBIO)},
  pages={1092--1097},
  year={2022},
  organization={IEEE}
}

@inproceedings{dantu2023adaptive,
  title={Adaptive anti-swing control for clasping operations in quadrotors with cable-suspended payload},
  author={Dantu, Swati and Yadav, Rishabh Dev and Rachakonda, Ananth and Roy, Spandan and Baldi, Simone},
  booktitle={2023 62nd IEEE Conference on Decision and Control (CDC)},
  pages={503--508},
  year={2023},
  organization={IEEE}
}

@inproceedings{ganguly2021efficient,
  title={Efficient manoeuvring of quadrotor under constrained space and predefined accuracy},
  author={Ganguly, Sourish and Sankaranarayanan, Viswa N and Suraj, BVSG and Yadav, Rishabh Dev and Roy, Spandan},
  booktitle={2021 IEEE/RSJ International Conference on Intelligent Robots and Systems (IROS)},
  pages={6352--6357},
  year={2021},
  organization={IEEE}
}

@inproceedings{gupta2024adaptive,
  title={Adaptive control of quadrotor under actuator loss and unknown state-dependent dynamics},
  author={Gupta, Saksham and Sharma, Amitabh and Mulgundkar, Aditya and Yadav, Rishabh Dev and Roy, Spandan},
  booktitle={2024 IEEE 20th international conference on automation science and engineering (CASE)},
  pages={717--722},
  year={2024},
  organization={IEEE}
}

@article{ganguly2022robust,
  title={Robust manoeuvring of quadrotor under full state constraints},
  author={Ganguly, Sourish and Sankaranarayanan, Viswa N and Suraj, BVSG and Yadav, Rishabh Dev and Roy, Spandan},
  journal={IFAC-PapersOnLine},
  volume={55},
  number={1},
  pages={32--37},
  year={2022},
  publisher={Elsevier}
}

@article{yadav2024modular,
  title={Modular adaptive aerial manipulation under unknown dynamic coupling forces},
  author={Yadav, Rishabh Dev and Dantu, Swati and Pan, Wei and Sun, Sihao and Roy, Spandan and Baldi, Simone},
  journal={IEEE/ASME Transactions on Mechatronics},
  volume={30},
  number={4},
  pages={2688--2698},
  year={2024},
  publisher={IEEE}
}

@article{dantu2024adaptive,
  title={Adaptive Tracking and Anti-Swing Control of Quadrotors Carrying Suspended Payload Under State-Dependent Uncertainty},
  author={Dantu, Swati and Yadav, Rishabh Dev and Rachakonda, Ananth and Roy, Spandan and Baldi, Simone},
  journal={IEEE/ASME Transactions on Mechatronics},
  year={2024},
  publisher={IEEE}
}

@article{yadav2025integrated,
  title={An integrated approach to aerial grasping: Combining a bistable gripper with adaptive control},
  author={Yadav, Rishabh Dev and Jones, Brycen and Gupta, Saksham and Sharma, Amitabh and Sun, Jiefeng and Zhao, Jianguo and Roy, Spandan},
  journal={IEEE/ASME Transactions on Mechatronics},
  year={2025},
  publisher={IEEE}
}

@inproceedings{sharma2025impedance,
  title={Impedance and Stability Targeted Adaptation for Aerial Manipulator with Unknown Coupling Dynamics},
  author={Sharma, Amitabh and Gupta, Saksham and Singh, Shivansh Pratap and Yadav, Rishabh Dev and Song, Hongyu and Pan, Wei and Roy, Spandan and Baldi, Simone},
  booktitle={2025 25th International Conference on Control, Automation and Systems (ICCAS)},
  pages={471--476},
  year={2025},
  organization={IEEE}
}
\end{document}